\documentclass[11pt]{article}
\usepackage[T1]{fontenc}
\usepackage[utf8]{inputenc}
\usepackage[margin=1in]{geometry}
\usepackage{amsmath,amssymb}
\usepackage{xcolor}
\usepackage{graphicx}
\usepackage{array}
\usepackage{needspace}
\usepackage{placeins}
\usepackage{float}
\usepackage{tikz}
\usepackage{pgfplots}
\pgfplotsset{compat=1.18}
\usepgfplotslibrary{groupplots}
\usetikzlibrary{arrows.meta,calc}
\usepackage{hyperref}
\hypersetup{hidelinks,breaklinks=true,pdfauthor={Ujjawal K. Panchal},pdftitle={FunDM: Cheap and Label-Free Decision Models from Frozen Language Models}}
\definecolor{fundmBlue}{RGB}{39,104,157}
\definecolor{fundmOrange}{RGB}{217,118,43}
\newcommand{\R}{\mathbb{R}}
\newcommand{\foh}{FunDM}
\newcommand{\fohc}{FunDM+1T}
\newcommand{\fohce}{FunDM-CE}
\newcommand{\fohcce}{FunDM-CE+1T}

\title{\textbf{\foh{}: Cheap \& Label-Free Decision Models from Frozen Language Models}}
\author{Ujjawal K. Panchal\\Fundamental AI\\\href{mailto:uj@fundamental.ai}{\texttt{uj@fundamental.ai}}}
\date{Preprint}
\begin{document}
\maketitle

\begin{abstract}
Decision models such as Jev score candidate answers or actions.
A small decision head trained label-free can recover about $99\%$
of a frozen language model's multiple-choice accuracy while
using only two-thirds of its transformer layers. It learns the full model's answer
probabilities with KL divergence, without correct-answer labels for training.
One backbone pass reads the question and
all options together; the head returns probabilities for every
option. We call this model \foh{} (Fun Decision Model).
Using $24$ of Qwen3-8B's $36$ layers and skipping the LM head,
its +1T variant reaches $75.0\%$ mean accuracy over eight benchmarks,
versus $75.7\%$ for the full model. On MMLU-Pro, both heads achieve
$1.56$ to $1.57\times$ batched vLLM throughput with about one-third
lower peak GPU allocator memory.
Both heads score higher than released Contrastive Language Models (CLM),
fine-tuned CLM, and CLM-shaped heads trained from scratch on the
eight multiple-choice benchmarks and
coding-agent selection. Separately trained agent heads learn judge
probabilities without pass/fail labels and select passing runs for
$74.4$ to $74.6$ of $100$ held-out tasks on average, versus $74$ for the
full-model judge and $52.0$ for a CLM-shaped action-trained head.
Code is available at: \url{https://github.com/Ujjawal-K-Panchal/FunDM}.
\end{abstract}

\section{Introduction}
\label{sec:intro}
A capable pretrained language model already knows how to choose
among answers, solutions, and agent runs. We reuse this capability
by training a small head on the frozen model's answer probabilities,
without retraining the backbone or collecting correct-answer labels
for head fitting.

Models that score candidate answers or actions given a question
or state are widely called \emph{Decision Models} (DMs). TypeSafe
AI's Jev \cite{almeida2026jev} and Contrastive Language Models (CLM)
\cite{kwok2026clm} are examples. \foh{} (Fun Decision Model)
reuses the decision capability of pretrained open-weights language models. Like CLM, it uses a frozen
Qwen3-8B backbone \cite{yang2025qwen3}.

Unlike CLM, \foh{} reads the question and all options together;
a single forward pass through the backbone and head produces a
score for each candidate option. Each option state sees the
question and preceding options; the final
answer state sees the whole list. Two learned projections score
the options. \fohc{} adds a bidirectional transformer layer so
earlier option states can attend to later ones inside the head.
Shared scoring weights support a variable number of options,
provided the prompt fits the context window (Figure~\ref{fig:overview}).

We cache full-model letter probabilities and backbone states, then
train the head by teacher-only knowledge distillation (KD).
Both \foh{} and \fohc{} use 24 of Qwen3-8B's 36 layers as a
fixed two-thirds early-exit depth. This tests how much
decision capability a backbone prefix retains while skipping
12 layers and the LM head. Section~\ref{sec:depth} examines
other budgets, including layer 32, which skips only four layers.

CLM encodes the question and each option separately, allowing
option vectors to be cached. Our supervised cross-entropy (CE)
baselines examine encoding and head architecture. Correct-answer
labels support evaluation, a
training-pool prompt comparison, and these ablations; they do not
fit the main FunDM heads or select their checkpoints.

Our contributions are:
\begin{itemize}
\item \textbf{Cheap, label-free decision training.}
Only a 1.05M- or 5.38M-parameter head is trained. Caching teacher
probabilities and states at five depths on 38k rows takes about
16 minutes on one DGX Spark; each head fits in under 30 seconds.
\item \textbf{Recovering pretrained decision capability.}
At 24 layers, \fohc{} retains 99.2\% of the teacher's mean
accuracy over eight benchmarks. Separately trained agent heads
show comparable selection performance; we report five-seed
results and paired uncertainty.
\item \textbf{Inference efficiency and component analysis.}
The 24-layer heads provide $1.56$--$1.57\times$ batched vLLM
throughput and about 33\% lower peak GPU allocator memory on
MMLU-Pro. Ablations examine training targets, encoding, head
architecture, exit depth, data scale, and prompt format.
\end{itemize}

\section{Related Work}
\label{sec:related}

\paragraph{Shared-Context Selection.}
Jina-reranker-v3 applies causal attention to the query and candidate
documents in one context, then projects their contextual final-token
states for ranking \cite{wang2025jina}.
CLM trains heads on separately encoded states
\cite{kwok2026clm}. LLM2Jev uses next-token probabilities of numeric
identifiers and studies listwise fine-tuning \cite{li2026llm2jev}.
We train small heads on option states with a frozen backbone and
test decisions before its final layer.

\paragraph{Distillation and Intermediate-Layer Readouts.}
Knowledge distillation trains a student on a teacher's probabilities
\cite{hinton2015distilling}. Tuned Lens uses KL to train affine
maps from intermediate states to the full model's next-token
predictions \cite{belrose2023tuned}. Our head scores a variable
number of options using their states and the answer-position state.
Probing work studies the gap between knowledge in hidden states
and model outputs \cite{park2025kappa,villuri2026probes}.

Early-exit methods include
DeeBERT, Depth-Adaptive Transformer, and CALM
\cite{xin2020deebert,elbayad2020depth,schuster2022calm}. LayerSkip
trains the backbone for early exits \cite{elhoushi2024layerskip}.
Pruning studies remove deeper layers
\cite{gromov2024deeper,men2024shortgpt}. We use a fixed exit depth
and train only the small choice head.

\paragraph{Judges and Verifiers.}
Verifiers rank sampled math solutions and coding-agent runs
\cite{cobbe2021verifiers,pan2024swegym}. Prompted language models
can also judge outputs directly \cite{zheng2023judge}. We distill
a prompted run judge into our option head without outcome labels.
The task is to select among existing runs; we do not change how
the runs are generated.

\section{Method}
\label{sec:method}

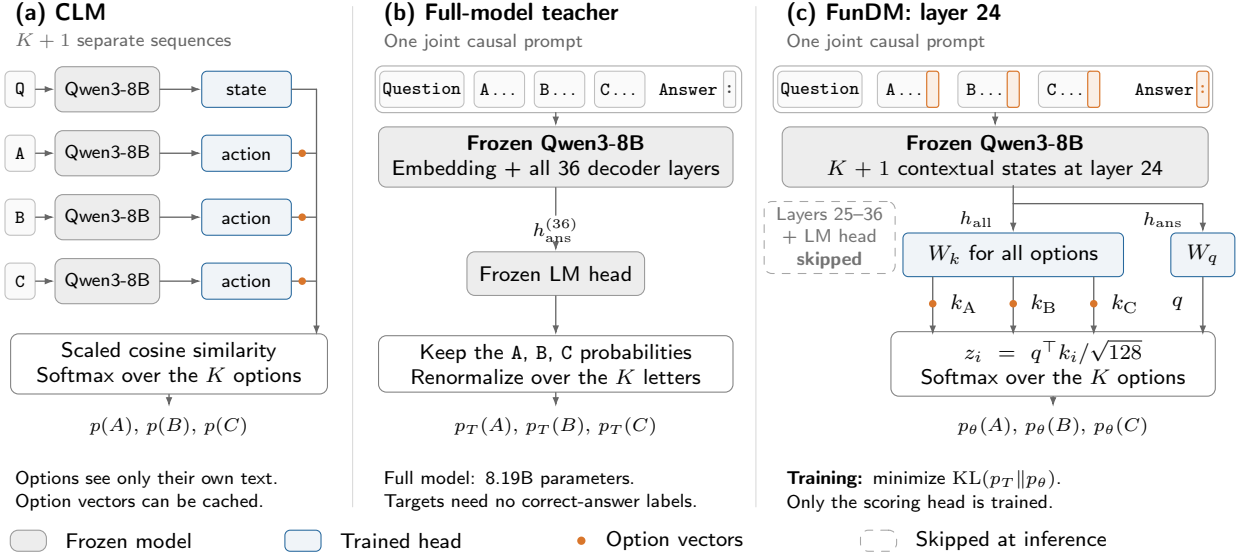
\begin{figure*}[!t]
\centering
\resizebox{\linewidth}{!}{%
\begin{tikzpicture}[
  x=1cm,y=1cm,
  font=\sffamily\fontsize{8}{9.5}\selectfont,
  line cap=round,line join=round,
  token/.style={draw=black!25,fill=black!2,rounded corners=2pt,
    minimum height=0.48cm,inner sep=1pt,align=center,
    font=\sffamily\fontsize{7}{8.5}\selectfont},
  frozen/.style={draw=black!40,fill=black!7,rounded corners=3pt,
    minimum height=0.76cm,inner sep=3pt,align=center},
  learned/.style={draw=fundmBlue,fill=fundmBlue!7,rounded corners=2pt,
    minimum height=0.58cm,inner sep=3pt,align=center},
  output/.style={draw=black!50,fill=white,rounded corners=3pt,
    minimum height=0.75cm,inner sep=3pt,align=center},
  flow/.style={-{Latex[length=1.3mm,width=0.9mm]},draw=black!60,line width=0.5pt},
  wire/.style={draw=black!60,line width=0.5pt},
  cue/.style={draw=black!25,fill=black!2,line width=0.45pt},
  skipped/.style={draw=black!35,text=black!55,dashed,rounded corners=3pt,
    fill=white,inner sep=3pt,align=center},
  heading/.style={anchor=west,font=\sffamily\fontsize{9.2}{11}\selectfont\bfseries},
  subtitle/.style={anchor=west,text=black!60,font=\sffamily\fontsize{7.5}{9}\selectfont},
  footer/.style={anchor=north west,align=left,font=\sffamily\fontsize{7.5}{9}\selectfont},
  probability/.style={font=\sffamily\fontsize{7.5}{9}\selectfont}
]
\path[use as bounding box] (0,-7.25) rectangle (16.46,0.20);
\draw[black!15,line width=0.4pt] (4.66,0.10) -- (4.66,-6.55);
\draw[black!15,line width=0.4pt] (10.02,0.10) -- (10.02,-6.55);

\node[heading] at (0.04,0) {(a) CLM};
\node[subtitle] at (0.04,-0.37) {$K+1$ separate sequences};
\foreach \yy/\sym/\projection in {
  -1.00/Q/state,
  -1.85/A/action,
  -2.70/B/action,
  -3.55/C/action}{
  \node[token,minimum width=0.40cm] (clm-input-\sym) at (0.24,\yy) {\texttt{\sym}};
  \node[frozen,text width=1.25cm,minimum height=0.60cm,inner sep=2pt,
    font=\sffamily\fontsize{7}{8.5}\selectfont]
    (clm-model-\sym) at (1.40,\yy) {Qwen3-8B};
  \node[learned,text width=1.05cm,minimum height=0.48cm,inner sep=2pt,
    font=\sffamily\fontsize{7.5}{9}\selectfont]
    (clm-map-\sym) at (3.25,\yy) {\projection};
  \draw[flow] (clm-input-\sym.east) -- (clm-model-\sym.west);
  \draw[flow] (clm-model-\sym.east) -- (clm-map-\sym.west);
  \draw[wire] (clm-map-\sym.east) -- (4.18,\yy);
  \ifdim\yy pt<-1.50pt
    \fill[fundmOrange] (3.99,\yy) circle (0.05);
  \fi
}
\draw[wire] (4.18,-1.00) -- (4.18,-4.10);
\node[output,text width=4.00cm] (clm-score) at (2.22,-4.65)
  {Scaled cosine similarity\\Softmax over the $K$ options};
\draw[flow] (4.18,-4.10) -- (4.18,-4.27);
\node[probability] (clm-probs) at (2.22,-5.48) {$p(A),\,p(B),\,p(C)$};
\draw[flow] (clm-score.south) -- (clm-probs.north);
\node[footer,text width=4.25cm] at (0.04,-5.93)
  {Options see only their own text.\\Option vectors can be cached.};

\node[heading] at (4.94,0) {(b) Full-model teacher};
\node[subtitle] at (4.94,-0.37) {One joint causal prompt};
\draw[draw=black!30,rounded corners=3pt,line width=0.45pt]
  (4.96,-1.31) rectangle (9.83,-0.69);
\node[token,text width=1.05cm] at (5.58,-1.00) {\texttt{Question}};
\foreach \x/\sym in {6.61/A,7.41/B,8.21/C}
  {\node[token,text width=0.60cm] at (\x,-1.00) {\texttt{\sym}\,\ldots{}};}
\node[font=\sffamily\fontsize{7}{8.5}\selectfont] at (9.12,-1.00) {\texttt{Answer}};
\draw[cue,rounded corners=1pt] (9.59,-1.24) rectangle (9.75,-0.76);
\node[text=black!75,font=\sffamily\fontsize{7}{8.5}\selectfont\ttfamily]
  at (9.67,-1.00) {:};
\node[frozen,text width=4.50cm] (teacher-model) at (7.36,-1.90)
  {\textbf{Frozen Qwen3-8B}\\Embedding + all 36 decoder layers};
\draw[flow] (7.36,-1.31) -- (teacher-model.north);
\node[probability,fill=white,inner sep=1pt] (teacher-state) at (7.36,-2.87)
  {$h_{\mathrm{ans}}^{(36)}$};
\draw[wire] (teacher-model.south) -- (teacher-state.north);
\node[frozen,text width=2.15cm,minimum height=0.58cm] (lm-head) at (7.36,-3.45)
  {Frozen LM head};
\draw[flow] (teacher-state.south) -- (lm-head.north);
\node[output,text width=4.50cm] (teacher-letters) at (7.36,-4.65)
  {Keep the \texttt{A}, \texttt{B}, \texttt{C} probabilities\\Renormalize over the $K$ letters};
\draw[flow] (lm-head.south) -- (teacher-letters.north);
\node[probability] (teacher-probs) at (7.36,-5.48) {$p_T(A),\,p_T(B),\,p_T(C)$};
\draw[flow] (teacher-letters.south) -- (teacher-probs.north);
\node[footer,text width=4.75cm] at (4.94,-5.93)
  {Full model: 8.19B parameters.\\Targets need no correct-answer labels.};

\node[heading] at (10.29,0) {(c) FunDM: layer 24};
\node[subtitle] at (10.29,-0.37) {One joint causal prompt};
\draw[draw=black!30,rounded corners=3pt,line width=0.45pt]
  (10.25,-1.31) rectangle (16.09,-0.69);
\node[token,text width=1.05cm] at (10.87,-1.00) {\texttt{Question}};
\foreach \x/\sym in {12.01/A,13.08/B,14.15/C}
  {\node[token,text width=0.68cm] at (\x,-1.00) {\texttt{\sym}\,\ldots{}};}
\node[font=\sffamily\fontsize{7}{8.5}\selectfont] at (15.45,-1.00) {\texttt{Answer}};
\foreach \x in {12.37,13.44,14.51,15.96}
  {\draw[cue,draw=fundmOrange,fill=fundmOrange!12,rounded corners=1pt] (\x-0.08,-1.24) rectangle (\x+0.08,-0.76);}
\node[text=fundmOrange,font=\sffamily\fontsize{7}{8.5}\selectfont\ttfamily]
  at (15.96,-1.00) {:};
\node[frozen,text width=5.45cm] (student-model) at (13.20,-1.90)
  {\textbf{Frozen Qwen3-8B}\\$K+1$ contextual states at layer 24};
\draw[flow] (13.20,-1.31) -- (student-model.north);
\node[skipped,text width=1.60cm,minimum height=0.88cm,inner sep=2pt,
  font=\sffamily\fontsize{7}{8.5}\selectfont] (skip) at (11.00,-2.97)
  {Layers 25--36\\+ LM head\\\textbf{skipped}};
\node[learned,text width=2.72cm] (wk) at (13.44,-3.20) {$W_k$ for all options};
\node[learned,text width=0.64cm] (wq) at (15.96,-3.20) {$W_q$};
\draw[wire] (13.44,-2.28) -- (13.44,-2.52);
\draw[flow] (13.44,-2.52) -- (wk.north);
\node[probability,anchor=east] at (13.31,-2.73) {$h_{\mathrm{all}}$};
\draw[flow] (13.44,-2.52) -- (15.96,-2.52) -- (wq.north);
\node[probability,anchor=east] at (15.83,-2.73) {$h_{\mathrm{ans}}$};
\node[output,text width=4.10cm] (student-score) at (13.97,-4.65)
  {$z_i=q^\top k_i/\sqrt{128}$\\Softmax over the $K$ options};
\foreach \x/\sym in {12.37/A,13.44/B,14.51/C} {
  \draw[flow] (wk.south -| \x,0) -- (\x,-4.27);
  \fill[fundmOrange] (\x,-3.85) circle (0.05);
  \node[anchor=west] at (\x+0.10,-3.85) {$k_{\mathrm{\sym}}$};
}
\draw[flow] (wq.south) -- (15.96,-4.27);
\node[anchor=east] at (15.83,-3.85) {$q$};
\node[probability] (student-probs) at (13.97,-5.48)
  {$p_\theta(A),\,p_\theta(B),\,p_\theta(C)$};
\draw[flow] (student-score.south) -- (student-probs.north);
\node[footer,text width=5.85cm] at (10.29,-5.93)
  {\textbf{Training:} minimize $\mathrm{KL}(p_T\|p_\theta)$.\\Only the scoring head is trained.};

\node[frozen,text width=0.48cm,minimum height=0.28cm,inner sep=0pt] at (0.35,-7.00) {};
\node[anchor=west] at (0.71,-7.00) {Frozen model};
\node[learned,text width=0.48cm,minimum height=0.28cm,inner sep=0pt] at (4.00,-7.00) {};
\node[anchor=west] at (4.36,-7.00) {Trained head};
\fill[fundmOrange] (7.70,-7.00) circle (0.055);
\node[anchor=west] at (7.89,-7.00) {Option vectors};
\node[skipped,text width=0.48cm,minimum height=0.28cm,inner sep=0pt] at (11.65,-7.00) {};
\node[anchor=west] at (12.01,-7.00) {Skipped at inference};
\end{tikzpicture}
}
\caption{\textbf{(a)} CLM uses separate encoding and scaled cosine
similarity; option vectors can be cached. \textbf{(b)} The frozen full
model supplies letter probabilities $p_T$. \textbf{(c)} \foh{} matches
$p_T$ with KL divergence using normalized joint states at layer 24.
$h_{\mathrm{all}}$ stacks option states for $W_k$; $h_{\mathrm{ans}}$
feeds $W_q$. Remaining layers and the LM head are skipped.
Optional \fohc{} mixes these $K+1$ readouts bidirectionally before
the projections; its transformer layer is not shown.}
\label{fig:overview}
\end{figure*}

\foh{} scores internal representations, called \emph{states}, from
a frozen backbone. Figure~\ref{fig:overview} contrasts its joint
encoding and teacher-only training with CLM's separate encoding.

\subsection{Joint Encoding}
For a question $x$ and $K$ options $\mathcal C=(c_1,\ldots,c_K)$,
we form one prompt containing the question, lettered options, and
an answer cue. At a chosen layer $L$, we take one state at each
option end and one at the answer position. The backbone is causal:
an option state sees the question and options up to that point,
while the answer state sees the whole prompt.

We normalize each state by its root mean square. Denote the option
states by $h_i^{(L)}\in\R^{4096}$ and the answer state by
$h_{\mathrm{ans}}^{(L)}$. These $K+1$ states are the head's only
inputs; $h_{\mathrm{all}}$ in Figure~\ref{fig:overview} denotes the
stack of option states. Appendix~\ref{sec:prompt} gives the prompt
and extraction details.

\subsection{Scoring Mechanism}
Two learned projections $W_q,W_k\in\R^{128\times4096}$ map
the answer state to a query $q$ and each option state to a key $k_i$:
\begin{equation}
q = W_q h_{\mathrm{ans}}^{(L)},\qquad
k_i = W_k h_i^{(L)}.
\end{equation}
A scaled dot product scores each option; softmax gives its probability:
\begin{equation}
\begin{gathered}
z_i = \frac{q^\top k_i}{\sqrt{128}},\\
p_\theta(i\mid x,\mathcal C) =
\frac{\exp z_i}{\sum_{j=1}^K\exp z_j}.
\end{gathered}
\end{equation}
The head produces one score per option, so increasing $K$ increases
the number of scores without changing the head's parameter count. We choose the
option with the highest probability. The base head has 1.05M
trainable parameters, while \fohc{} has 5.38M (5{,}383{,}168).

\fohc{} first maps all $K+1$ states to 512 dimensions, then mixes
them with one bidirectional transformer layer. Each selected state
can attend to all the others, while the backbone remains causal.
The updated option and answer states use the same scoring rule,
with $W_q,W_k\in\R^{128\times512}$ (Appendix~\ref{app:impl}).

\subsection{Head Training}
\paragraph{Teacher-Probability Training.}
The teacher is the same frozen backbone run at full depth with its
original language-model head. It shares the head's multi-line,
lettered prompt, which asks for an option letter.
Appendix~\ref{sec:prompt} gives this format and the validation
comparison that motivated its choice.

At the answer position, the teacher provides next-token
log-probabilities $\ell_i$ for the option letters. We normalize over
these $K$ letters to obtain $p_T$ and train the head to match it:
\begin{align}
p_T(i\mid x,\mathcal C)&=\frac{\exp\ell_i}{\sum_{j=1}^K\exp\ell_j},\\
\mathcal L_{\mathrm{KL}}&=\sum_i p_T(i\mid x,\mathcal C)
\log\frac{p_T(i\mid x,\mathcal C)}{p_\theta(i\mid x,\mathcal C)}.
\end{align}
This teacher-only distillation retains probabilities for all choices
\cite{hinton2015distilling}, with no CE term against correct-answer
labels. For a fixed teacher, it has the same gradients as
cross-entropy with soft targets. One full backbone pass per MCQ
row caches probabilities and states at all tested layers; only
the head is updated thereafter. Validation KL selects the
checkpoint (Section~\ref{sec:setup}).

\paragraph{Inference Procedure.}
\foh{} runs backbone layers 1 through $L$ ($L=24$ in our case), then applies the head
to produce option probabilities and selects the most probable
option. The remaining backbone layers and language-model head
are skipped.

\section{Experimental Setup}
\label{sec:setup}
All experiments use a single NVIDIA DGX Spark (GB10 GPU).
Code and reproduction scripts are available at:
\url{https://github.com/Ujjawal-K-Panchal/FunDM}.

\paragraph{Data and Checkpoint Selection.}
Our 38{,}067-row MCQ pool uses ARC, SciQ, and MedMCQA
\cite{clark2018arc,welbl2017sciq,pal2022medmcqa}. About half the
rows add distractors from other questions, giving up to ten
options. A question-grouped split gives 36{,}164 training and
1{,}903 validation rows. Validation KL selects FunDM checkpoints,
without correct-answer labels for fitting or selection. Results
use seed 0 unless stated. Appendix~\ref{app:impl} gives fitting,
prompt, and data details.

\paragraph{Baselines.}
The full-model reference uses all 36 Qwen3-8B layers and its LM
head to choose an option letter. Released CLM-v0.1-8B, fine-tuned
CLM, and CLM-shaped heads trained from scratch all use the same
frozen backbone. Released and fine-tuned CLM use the authors' input
formats, encoding the question and each option separately
(Appendix~\ref{app:clm}). The from-scratch controls use separate encoding
and CE on our rows (Section~\ref{sec:grid}).
For fine-tuning, the authors' script updates released CLM's two
MLP heads and learned cosine scale with bidirectional in-batch
InfoNCE. Its random 90/10 row split gives 34{,}260 fitting rows;
validation accuracy selects the checkpoint. We average seeds
0--2. Appendix~\ref{app:clm} details comparison and reproduction limits.

We also evaluate \fohce{} and \fohcce{}, supervised variants of
the base and +1T head architectures trained with CE on correct-answer
labels. Their original MCQ recipe uses all 36 layers and a single-line
prompt (Appendix~\ref{app:ce-details}).

\paragraph{Evaluation.}
We evaluate non-thinking option-letter selection on MMLU-Pro's
12{,}032 questions and seven other benchmarks: MMLU, OpenBookQA,
CommonsenseQA, PIQA, HellaSwag, WinoGrande, and BoolQ
\cite{wang2024mmlupro,hendrycks2020mmlu,mihaylov2018obqa,
talmor2019csqa,bisk2020piqa,zellers2019hellaswag,
sakaguchi2020winogrande,clark2019boolq}.
We report accuracy and an equally weighted eight-benchmark mean.
The main MCQ and agent heads use the same fixed layer-24 budget;
parameter and decoder-layer savings are architectural estimates.

\section{Results}
\label{sec:results}
\subsection{Multiple-Choice Performance}
\label{sec:sets}
Both FunDM variants outperform released, fine-tuned, and
from-scratch CLM-shaped heads on every benchmark
(Table~\ref{tab:sets}). Their mean accuracies exceed fine-tuned
CLM's by 28.3 and 28.6 percentage points. At 24 layers, \foh{}
and \fohc{} score 0.747 and 0.750 versus the teacher's 0.757,
retaining 98.7\% and 99.2\% of its mean accuracy. The teacher
itself already exceeds fine-tuned CLM's 0.464.

\begin{table*}[!t]
\centering\footnotesize
\setlength{\tabcolsep}{3pt}
\begin{tabular}{lccccc|c}
\hline
Set & CLM & CLM ft. & CLM-shaped & \foh{} & \multicolumn{1}{c|}{\fohc{}} & Teacher \\
\hline
MMLU-Pro & 0.180 & 0.182 & 0.157 & \underline{0.426} & \textbf{0.443} & 0.451 \\
MMLU & 0.358 & 0.368 & 0.299 & \underline{0.684} & \textbf{0.685} & 0.706 \\
OpenBookQA & 0.338 & \underline{0.386} & 0.288 & \textbf{0.834} & \textbf{0.834} & 0.856 \\
CommonsenseQA & 0.389 & 0.426 & 0.237 & \textbf{0.799} & \underline{0.796} & 0.792 \\
PIQA & 0.609 & 0.610 & 0.595 & \underline{0.849} & \textbf{0.856} & 0.854 \\
HellaSwag & 0.423 & 0.423 & 0.399 & \underline{0.833} & \textbf{0.840} & 0.827 \\
WinoGrande & 0.482 & 0.499 & 0.505 & \textbf{0.682} & \underline{0.680} & 0.696 \\
BoolQ & 0.764 & 0.819 & 0.378 & \underline{0.865} & \textbf{0.869} & 0.870 \\
\hline
Mean & 0.443 & 0.464 & 0.357 & \underline{0.747} & \textbf{0.750} & 0.757 \\
\hline
Total parameters & 7.59B & 7.59B & 7.59B & 5.25B & 5.26B & 8.19B \\
Trainable parameters & 18.9M & 18.9M & 18.9M & 1.05M & 5.38M & 0 \\
\hline
Layers run (of 36) & 36 & 36 & 36 & 24 & 24 & 36 + LM head \\
\hline
\end{tabular}
\caption{Eight-benchmark accuracy. CLM ft. fine-tunes released CLM
on our rows (three-seed mean); CLM-shaped trains its architecture
from scratch on our rows with separate encoding (Table~\ref{tab:grid}). CLM-shaped
and FunDM heads use seed 0. The teacher uses full-model letter probabilities.
Total parameters include the inference backbone and head.
Bold/underlining mark the best/second-best non-teacher scores;
ties share a rank.}
\label{tab:sets}
\end{table*}

Both heads are within 2.5 percentage points of the teacher on
each benchmark. Lower negative log-likelihood (NLL) means more
probability on correct answers; \fohc{} has lower NLL on six of
eight sets, including MMLU-Pro (1.98 versus 2.22). Similar accuracy
does not imply identical answers: mean teacher agreement is about
88\%, and MMLU-Pro agreement is about 75\%.
Section~\ref{sec:ablations} examines
training targets and supervised encoding.

\subsection{Inference Efficiency}
\label{sec:efficiency}
We compare six models in vLLM on the same 500 MMLU-Pro questions,
with up to ten options each. All return option probabilities;
the full model uses LM-head letter probabilities.
Table~\ref{tab:efficiency} measures batched throughput,
batch-one first-decision latency, and peak GPU allocator memory.
The 24-layer heads achieve $1.56$--$1.57\times$ throughput,
reduce latency from 93.7 to 57.6--58.3 ms, and reduce peak memory
from 20.5 to 13.8 GiB (about 33\%).

\begin{table*}[!t]
\centering\footnotesize
\setlength{\tabcolsep}{4pt}
\begin{tabular}{lcccccc}
\hline
Model & Layers & Params & Peak (GiB) & Questions/s & First decision (ms) & Speedup \\
\hline
CLM (released) & 36 & 7.59B & 19.3 & 16.8 & 216.7 & $0.69\times$ \\
Full model + LM head & 36 & 8.19B & 20.5 & 24.5 & 93.7 & $1.00\times$ \\
\foh{} & 24 & 5.25B & 13.8 & 38.4 & 57.6 & $1.57\times$ \\
\fohc{} & 24 & 5.26B & 13.8 & 38.2 & 58.3 & $1.56\times$ \\
\fohce{} & 36 & 7.57B & 19.8 & 29.5 & 84.5 & $1.20\times$ \\
\fohcce{} & 36 & 7.57B & 19.6 & 29.5 & 86.9 & $1.20\times$ \\
\hline
\end{tabular}
\caption{BF16 vLLM inference on 500 MMLU-Pro questions with a
32{,}768-token KV pool: 16 active sequences for throughput,
batch-one requests for latency. Speedup is relative to the full model.
CE denotes training with correct-answer labels.
Peak is whole-workload CUDA allocator reserved memory, excluding non-PyTorch
allocations/context. Teacher/FunDM share a prompt; CE retains its
original prompt (Appendix~\ref{app:timing}).}
\label{tab:efficiency}
\end{table*}

These gains combine one joint backbone pass, a compact readout,
and reduced depth. Shared weights score a variable number of
options without increasing head parameters. Scoring only supplied
candidates replaces the full-vocabulary LM head; skipping it and
12 layers reduces total inference parameters by about 36\%.
CLM also supports a variable number of options, but encodes them
separately. With caches initially empty on each pass, this workload
has 0\% vector-cache and 0.82\% prefix-cache hits.

A separate PyTorch experiment gives 24-layer \foh{}
$1.50\times$ the teacher's throughput across eight benchmarks.
Preloading all 500 MMLU-Pro option vectors raises CLM from 14.1
to 31.7 questions/s ($2.25\times$), above the teacher (14.0) and
\foh{} (21.0) in that run. This assumes known candidates and
excludes their encoding cost; it is separate from vLLM
(Appendix~\ref{app:timing}).

\section{Agentic Selection}
\label{sec:agentic}
We distill a judge that selects among existing coding-agent runs.
This \emph{best-of-$N$ selection} is useful when an agent generates
several solutions but must return one, as in CodeMonkeys
\cite{ehrlich2025codemonkeys}.

CoderForge-Preview records coding attempts and test outcomes
\cite{ariyak2026coderforge}. We use its SWE-rebench subset of
GitHub issues, repository environments, and tests
\cite{badertdinov2025swerebench}. Unlike the fixed SWE-bench
set, SWE-rebench also maintains a rolling benchmark of fresh
issues \cite{jimenez2024swebench,badertdinov2025swerebench}.
Qwen3-Coder-480B generated these runs using OpenHands, a framework
for inspecting files, editing code, and executing commands
\cite{ariyak2026coderforge,wang2024openhands}; each run records
actions and observations from one attempt.

We select one run for each of 100 held-out tasks with 799 runs:
eight per task, except one with seven. Every task has passing and
failing runs. Random selection solves 54.9 tasks in expectation;
a perfect selector solves 100.

\paragraph{Judge Distillation.}
The full model reads the task and recent run text and predicts
whether the final changes will pass hidden tests. Normalizing
yes/no probabilities gives $p_T(\mathrm{yes})$; selecting the
highest solves 74 tasks. The student reads the same task and run
with lettered yes/no options, learns the judge probabilities with
KL, and ranks runs by its yes-minus-no logit. Pass/fail labels only
evaluate these heads. Appendix~\ref{app:agentic} specifies prompts
and text limits. Action-trained CLM baselines instead score
actions against preceding run text and aggregate scores.

\paragraph{Training and Performance.}
From 1{,}600 runs on 200 training tasks, each seed reserves ten
tasks for validation; validation KL selects the checkpoint.
Table~\ref{tab:agent} reports five-seed results at layer 24.
AUC measures how often a passing run outscores a failing run
within a task. The 5.25B-parameter backbone prefix uses 67\% of
full-model decoder-layer operations for equal-length inputs.

\begin{table*}[!t]
\centering\small
\setlength{\tabcolsep}{4pt}
\begin{tabular}{lcccc}
\hline
Selector & Total params & Head params & Tasks solved & AUC \\
\hline
Released CLM, no training$^\dagger$ & 7.59B & 18.9M & 44.0 & 0.458 \\
Released CLM, fine-tuned$^\dagger$ & 7.59B & 18.9M & $58.2\pm3.4$ & $0.581\pm0.008$ \\
CLM-shaped, action-trained & 7.59B & 18.9M & $52.0\pm4.2$ & $0.551\pm0.043$ \\
\foh{}, layer 24 & 5.25B & 1.05M & $\mathbf{74.6\pm1.5}$ & $\mathbf{0.765\pm0.025}$ \\
\fohc{}, layer 24 & 5.26B & 5.38M & $\underline{74.4\pm2.2}$ & $\underline{0.748\pm0.022}$ \\
\hline
Full-model judge (teacher) & 8.19B & 0 & 74 & 0.760 \\
\hline
\end{tabular}
\caption{CoderForge selection on 100 tasks; AUC is within-task.
Trained heads report five-seed means $\pm$ sample standard deviations.
Total parameters include the backbone and head. Bold/underlining
mark the best/second-best non-teacher means.
$^\dagger$Released CLM uses its authors' mean-cosine scoring rule,
before/after fine-tuning on 200 training tasks; random choice solves 54.9 tasks.}
\label{tab:agent}
\end{table*}

Both heads select passing runs at rates comparable to the judge,
without establishing superiority or statistical equivalence.
Their 95\% paired task-bootstrap intervals for differences include
zero; for \fohc{}, they span $-8.0$ to $+8.8$ tasks and
$-0.059$ to $+0.033$ AUC. Heads choose different runs from the
judge on 60--69 tasks. Seed deviations in the table do not measure
task-sampling uncertainty.

The judge already solves 74 tasks versus the action-trained
CLM-shaped head's 52.0. Released CLM before/after fine-tuning
solves 44.0/58.2 using its authors' mean-cosine scoring rule. These comparisons change
inputs, supervision, and run scoring. Section~\ref{sec:depth} examines agent
exit depth; Appendix~\ref{app:deepswe} tests CE heads on released
separate DeepSWE embeddings.

\section{Ablations}
\label{sec:ablations}
We examine targets, encoding, head architecture, exit depth,
data, and prompts. The matched target study uses the layer-24
FunDM recipe; encoding and architecture studies use the original
final-layer CE recipe. Appendix~\ref{app:ablation-details} gives
supporting diagnostics and Appendix~\ref{sec:prompt} covers prompts.

\subsection{Effect of the Training Target}
\label{sec:target}
Table~\ref{tab:target} fixes the prompt, layer-24 states, head,
and fitting, changing only targets. Each run selects its checkpoint
by validation loss under its target. Full teacher probabilities
give the highest observed mean accuracy: 0.747/0.751 for the two
heads, versus 0.735/0.740 for teacher top-choice labels and
0.713/0.734 for correct-option labels.

\begin{table*}[!t]
\centering\small
\begin{tabular}{lcc}
\hline
Target (layer 24, FunDM prompt) & \foh{} & \fohc{} \\
\hline
Correct-option labels & $0.713\pm0.016$ & $0.734\pm0.001$ \\
Teacher's top choice & $\underline{0.735\pm0.006}$ & $\underline{0.740\pm0.003}$ \\
Teacher probabilities (FunDM) & $\mathbf{0.747\pm0.001}$ & $\mathbf{0.751\pm0.002}$ \\
\hline
Full model (teacher), 36 layers & \multicolumn{2}{c}{0.757} \\
\hline
\end{tabular}
\caption{Mean accuracy over eight benchmarks when only the training
target changes; states, prompt, head, and fitting are fixed. Means
$\pm$ sample standard deviations over seeds 0--2. Bold marks the best
target per column; underlining, the second best.}
\label{tab:target}
\end{table*}

\paragraph{Supervised CE Recipe.}
\label{sec:ce}
Our original \fohce{} and \fohcce{} baselines use all 36 layers,
a single-line \texttt{::}-delimited prompt, final-layer normalized
states, and one-hot correct-option labels:
\begin{equation}
\mathcal L_{\mathrm{CE}}=-\log p_\theta(y\mid x,\mathcal C).
\end{equation}
Validation NLL selects checkpoints. Their prompt, readout,
normalization, and depth differ from FunDM, so that gap does not
isolate the target. Appendix~\ref{app:ce-details} retains the full
MCQ results, head diagram, and action-trained agent results.

\subsection{Encoding Strategies under Cross-Entropy Training}
\label{sec:grid}
Table~\ref{tab:grid} compares option-only, question-plus-option,
and full-list encoding under CE. Joint reading improves mean
MMLU-Pro accuracy over option-only by 15.7, 17.3, and 16.4
percentage points for \fohce{}, \fohcce{}, and CLM-shaped MLPs.
Question plus option scores higher still (0.342/0.366/0.369
versus 0.317/0.332/0.327). Question context supplies the main
accuracy gain; joint reading saves computation. A joint MMLU-Pro
list uses 180 tokens per question versus 10.5 passes and 842
tokens for question plus option, which repeatedly reads the question.

\begin{table*}[!t]
\centering\small
\setlength{\tabcolsep}{4pt}
\begin{tabular}{lcccc}
\hline
Head architecture & \shortstack{Head\\params} & Option only (CLM format) & Question + option & Full list \\
\hline
\fohce{} & 1.05M & $0.160\pm0.000$ & $\mathbf{0.342\pm0.016}$ & $\underline{0.317\pm0.015}$ \\
\fohcce{} & 5.38M & $0.159\pm0.003$ & $\mathbf{0.366\pm0.004}$ & $\underline{0.332\pm0.001}$ \\
CLM-shaped MLPs & 18.9M & $0.163\pm0.005$ & $\mathbf{0.369\pm0.001}$ & $\underline{0.327\pm0.002}$ \\
\hline
Released CLM (pretrained) & 18.9M & 0.180 & -- & -- \\
\hline
\end{tabular}
\caption{MMLU-Pro CE accuracy, means $\pm$ sample standard deviations
over seeds 0--2. Option only: separate question and option embeddings
from CLM's vLLM path.
Question + option: each pair in its own sequence. Full list: the
question and all options together. The last two read final-layer
option-delimiter states. Bold/underlining mark the best/second-best
accuracy per row. Released CLM is an externally trained reference.}
\label{tab:grid}
\end{table*}

Question plus option and full list share extraction and readout;
option only uses CLM's. \fohcce{} mixes options inside its head
in every column. These CE results do not establish the same
encoding effect under teacher-only training.

\paragraph{Head Architecture.}
With identical joint CE states, \fohcce{} has the highest
single-seed accuracy, while CLM-shaped MLPs have the lowest NLL
and expected calibration error (ECE; confidence--accuracy mismatch).
Small accuracy differences should be read alongside seed variation
(Table~\ref{tab:gold-heads}, Appendix~\ref{app:architecture-details}).

\subsection{Effect of Exit Depth}
\label{sec:depth}
\paragraph{Multiple-Choice Selection.}
Table~\ref{tab:depth} shows the accuracy--budget tradeoff.
At 28 layers, mean accuracies reach 0.756/0.757; at 20 and 16,
heads retain 94--95\% and 75--76\% of teacher accuracy. The heads
differ by at most one percentage point at any tested depth.
Later validation KL need not improve test accuracy
(Appendix~\ref{app:kl-details}).

\begin{table*}[!t]
\centering\small
\setlength{\tabcolsep}{3pt}
\begin{tabular}{lcccc}
\hline
 & & & \multicolumn{2}{c}{Mean of 8 sets $\uparrow$} \\
Layers run (of 36) & Parameters & Layers used & \foh{} & \fohc{} \\
\hline
16 & 3.71B (45\%) & 44\% & 0.579 & 0.569 \\
20 & 4.48B (55\%) & 56\% & 0.708 & 0.716 \\
24 (two-thirds) & 5.25B (64\%) & 67\% & 0.747 & 0.750 \\
28 & 6.02B (74\%) & 78\% & 0.756 & 0.757 \\
32 & 6.80B (83\%) & 89\% & 0.753 & 0.755 \\
\hline
36 + LM head (teacher) & 8.19B (100\%) & 100\% & \multicolumn{2}{c}{0.757} \\
\hline
\end{tabular}
\caption{Equal-weight eight-benchmark accuracy by exit depth (seed 0).
Student counts include the embedding and retained decoder layers,
excluding scoring heads; the teacher includes its LM head.
Parameter and layer percentages are relative to the full model.}
\label{tab:depth}
\end{table*}

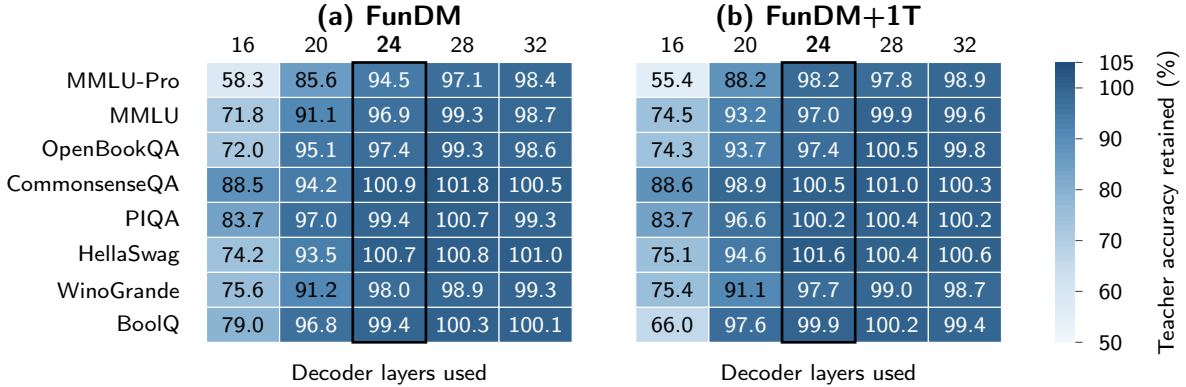
\begin{figure*}[!t]
\centering
\resizebox{\linewidth}{!}{%
\begin{tikzpicture}[x=1cm,y=1cm,font=\sffamily\fontsize{8.5}{10}\selectfont]
\path[use as bounding box] (-0.35,-0.67) rectangle (15.35,4.38);
\node[font=\sffamily\bfseries\fontsize{10}{12}\selectfont] at (4.90,4.08) {(a) \foh{}};
\node[] at (3.06,3.75) {16};
\node[] at (3.98,3.75) {20};
\node[font=\sffamily\bfseries\fontsize{8.5}{10}\selectfont] at (4.90,3.75) {24};
\node[] at (5.82,3.75) {28};
\node[] at (6.74,3.75) {32};
\node[anchor=east] at (2.40,3.30) {MMLU-Pro};
\definecolor{fundmHM000}{rgb}{0.85732,0.91255,0.95674}
\filldraw[fill=fundmHM000,draw=white,line width=0.4pt] (2.60,3.08) rectangle (3.52,3.52);
\node[text=black] at (3.06,3.30) {58.3};
\definecolor{fundmHM001}{rgb}{0.42295,0.61324,0.76354}
\filldraw[fill=fundmHM001,draw=white,line width=0.4pt] (3.52,3.08) rectangle (4.44,3.52);
\node[text=black] at (3.98,3.30) {85.6};
\definecolor{fundmHM002}{rgb}{0.27336,0.48136,0.65403}
\filldraw[fill=fundmHM002,draw=white,line width=0.4pt] (4.44,3.08) rectangle (5.36,3.52);
\node[text=white] at (4.90,3.30) {94.5};
\definecolor{fundmHM003}{rgb}{0.23465,0.43685,0.60759}
\filldraw[fill=fundmHM003,draw=white,line width=0.4pt] (5.36,3.08) rectangle (6.28,3.52);
\node[text=white] at (5.82,3.30) {97.1};
\definecolor{fundmHM004}{rgb}{0.21530,0.41460,0.58436}
\filldraw[fill=fundmHM004,draw=white,line width=0.4pt] (6.28,3.08) rectangle (7.20,3.52);
\node[text=white] at (6.74,3.30) {98.4};
\node[anchor=east] at (2.40,2.86) {MMLU};
\definecolor{fundmHM010}{rgb}{0.66513,0.78704,0.87929}
\filldraw[fill=fundmHM010,draw=white,line width=0.4pt] (2.60,2.64) rectangle (3.52,3.08);
\node[text=black] at (3.06,2.86) {71.8};
\definecolor{fundmHM011}{rgb}{0.32315,0.53740,0.71164}
\filldraw[fill=fundmHM011,draw=white,line width=0.4pt] (3.52,2.64) rectangle (4.44,3.08);
\node[text=black] at (3.98,2.86) {91.1};
\definecolor{fundmHM012}{rgb}{0.23805,0.44076,0.61166}
\filldraw[fill=fundmHM012,draw=white,line width=0.4pt] (4.44,2.64) rectangle (5.36,3.08);
\node[text=white] at (4.90,2.86) {96.9};
\definecolor{fundmHM013}{rgb}{0.20303,0.40048,0.56963}
\filldraw[fill=fundmHM013,draw=white,line width=0.4pt] (5.36,2.64) rectangle (6.28,3.08);
\node[text=white] at (5.82,2.86) {99.3};
\definecolor{fundmHM014}{rgb}{0.21127,0.40996,0.57952}
\filldraw[fill=fundmHM014,draw=white,line width=0.4pt] (6.28,2.64) rectangle (7.20,3.08);
\node[text=white] at (6.74,2.86) {98.7};
\node[anchor=east] at (2.40,2.42) {OpenBookQA};
\definecolor{fundmHM020}{rgb}{0.66263,0.78541,0.87826}
\filldraw[fill=fundmHM020,draw=white,line width=0.4pt] (2.60,2.20) rectangle (3.52,2.64);
\node[text=black] at (3.06,2.42) {72.0};
\definecolor{fundmHM021}{rgb}{0.26410,0.47071,0.64291}
\filldraw[fill=fundmHM021,draw=white,line width=0.4pt] (3.52,2.20) rectangle (4.44,2.64);
\node[text=white] at (3.98,2.42) {95.1};
\definecolor{fundmHM022}{rgb}{0.23011,0.43163,0.60213}
\filldraw[fill=fundmHM022,draw=white,line width=0.4pt] (4.44,2.20) rectangle (5.36,2.64);
\node[text=white] at (4.90,2.42) {97.4};
\definecolor{fundmHM023}{rgb}{0.20292,0.40036,0.56951}
\filldraw[fill=fundmHM023,draw=white,line width=0.4pt] (5.36,2.20) rectangle (6.28,2.64);
\node[text=white] at (5.82,2.42) {99.3};
\definecolor{fundmHM024}{rgb}{0.21312,0.41209,0.58174}
\filldraw[fill=fundmHM024,draw=white,line width=0.4pt] (6.28,2.20) rectangle (7.20,2.64);
\node[text=white] at (6.74,2.42) {98.6};
\node[anchor=east] at (2.40,1.98) {CommonsenseQA};
\definecolor{fundmHM030}{rgb}{0.36982,0.57286,0.73590}
\filldraw[fill=fundmHM030,draw=white,line width=0.4pt] (2.60,1.76) rectangle (3.52,2.20);
\node[text=black] at (3.06,1.98) {88.5};
\definecolor{fundmHM031}{rgb}{0.27721,0.48579,0.65865}
\filldraw[fill=fundmHM031,draw=white,line width=0.4pt] (3.52,1.76) rectangle (4.44,2.20);
\node[text=white] at (3.98,1.98) {94.2};
\definecolor{fundmHM032}{rgb}{0.17987,0.37385,0.54185}
\filldraw[fill=fundmHM032,draw=white,line width=0.4pt] (4.44,1.76) rectangle (5.36,2.20);
\node[text=white] at (4.90,1.98) {100.9};
\definecolor{fundmHM033}{rgb}{0.16702,0.35907,0.52642}
\filldraw[fill=fundmHM033,draw=white,line width=0.4pt] (5.36,1.76) rectangle (6.28,2.20);
\node[text=white] at (5.82,1.98) {101.8};
\definecolor{fundmHM034}{rgb}{0.18538,0.38019,0.54846}
\filldraw[fill=fundmHM034,draw=white,line width=0.4pt] (6.28,1.76) rectangle (7.20,2.20);
\node[text=white] at (6.74,1.98) {100.5};
\node[anchor=east] at (2.40,1.54) {PIQA};
\definecolor{fundmHM040}{rgb}{0.45684,0.63900,0.78116}
\filldraw[fill=fundmHM040,draw=white,line width=0.4pt] (2.60,1.32) rectangle (3.52,1.76);
\node[text=black] at (3.06,1.54) {83.7};
\definecolor{fundmHM041}{rgb}{0.23701,0.43956,0.61041}
\filldraw[fill=fundmHM041,draw=white,line width=0.4pt] (3.52,1.32) rectangle (4.44,1.76);
\node[text=white] at (3.98,1.54) {97.0};
\definecolor{fundmHM042}{rgb}{0.20124,0.39843,0.56749}
\filldraw[fill=fundmHM042,draw=white,line width=0.4pt] (4.44,1.32) rectangle (5.36,1.76);
\node[text=white] at (4.90,1.54) {99.4};
\definecolor{fundmHM043}{rgb}{0.18251,0.37688,0.54501}
\filldraw[fill=fundmHM043,draw=white,line width=0.4pt] (5.36,1.32) rectangle (6.28,1.76);
\node[text=white] at (5.82,1.54) {100.7};
\definecolor{fundmHM044}{rgb}{0.20295,0.40039,0.56954}
\filldraw[fill=fundmHM044,draw=white,line width=0.4pt] (6.28,1.32) rectangle (7.20,1.76);
\node[text=white] at (6.74,1.54) {99.3};
\node[anchor=east] at (2.40,1.10) {HellaSwag};
\definecolor{fundmHM050}{rgb}{0.62446,0.76052,0.86249}
\filldraw[fill=fundmHM050,draw=white,line width=0.4pt] (2.60,0.88) rectangle (3.52,1.32);
\node[text=black] at (3.06,1.10) {74.2};
\definecolor{fundmHM051}{rgb}{0.28770,0.49786,0.67124}
\filldraw[fill=fundmHM051,draw=white,line width=0.4pt] (3.52,0.88) rectangle (4.44,1.32);
\node[text=white] at (3.98,1.10) {93.5};
\definecolor{fundmHM052}{rgb}{0.18217,0.37650,0.54461}
\filldraw[fill=fundmHM052,draw=white,line width=0.4pt] (4.44,0.88) rectangle (5.36,1.32);
\node[text=white] at (4.90,1.10) {100.7};
\definecolor{fundmHM053}{rgb}{0.18042,0.37448,0.54250}
\filldraw[fill=fundmHM053,draw=white,line width=0.4pt] (5.36,0.88) rectangle (6.28,1.32);
\node[text=white] at (5.82,1.10) {100.8};
\definecolor{fundmHM054}{rgb}{0.17866,0.37246,0.54039}
\filldraw[fill=fundmHM054,draw=white,line width=0.4pt] (6.28,0.88) rectangle (7.20,1.32);
\node[text=white] at (6.74,1.10) {101.0};
\node[anchor=east] at (2.40,0.66) {WinoGrande};
\definecolor{fundmHM060}{rgb}{0.60220,0.74600,0.85330}
\filldraw[fill=fundmHM060,draw=white,line width=0.4pt] (2.60,0.44) rectangle (3.52,0.88);
\node[text=black] at (3.06,0.66) {75.6};
\definecolor{fundmHM061}{rgb}{0.32026,0.53520,0.71014}
\filldraw[fill=fundmHM061,draw=white,line width=0.4pt] (3.52,0.44) rectangle (4.44,0.88);
\node[text=black] at (3.98,0.66) {91.2};
\definecolor{fundmHM062}{rgb}{0.22199,0.42228,0.59238}
\filldraw[fill=fundmHM062,draw=white,line width=0.4pt] (4.44,0.44) rectangle (5.36,0.88);
\node[text=white] at (4.90,0.66) {98.0};
\definecolor{fundmHM063}{rgb}{0.20945,0.40786,0.57734}
\filldraw[fill=fundmHM063,draw=white,line width=0.4pt] (5.36,0.44) rectangle (6.28,0.88);
\node[text=white] at (5.82,0.66) {98.9};
\definecolor{fundmHM064}{rgb}{0.20318,0.40065,0.56981}
\filldraw[fill=fundmHM064,draw=white,line width=0.4pt] (6.28,0.44) rectangle (7.20,0.88);
\node[text=white] at (6.74,0.66) {99.3};
\node[anchor=east] at (2.40,0.22) {BoolQ};
\definecolor{fundmHM070}{rgb}{0.54335,0.70475,0.82614}
\filldraw[fill=fundmHM070,draw=white,line width=0.4pt] (2.60,0.00) rectangle (3.52,0.44);
\node[text=black] at (3.06,0.22) {79.0};
\definecolor{fundmHM071}{rgb}{0.23954,0.44247,0.61345}
\filldraw[fill=fundmHM071,draw=white,line width=0.4pt] (3.52,0.00) rectangle (4.44,0.44);
\node[text=white] at (3.98,0.22) {96.8};
\definecolor{fundmHM072}{rgb}{0.20109,0.39825,0.56730}
\filldraw[fill=fundmHM072,draw=white,line width=0.4pt] (4.44,0.00) rectangle (5.36,0.44);
\node[text=white] at (4.90,0.22) {99.4};
\definecolor{fundmHM073}{rgb}{0.18771,0.38287,0.55125}
\filldraw[fill=fundmHM073,draw=white,line width=0.4pt] (5.36,0.00) rectangle (6.28,0.44);
\node[text=white] at (5.82,0.22) {100.3};
\definecolor{fundmHM074}{rgb}{0.19106,0.38671,0.55527}
\filldraw[fill=fundmHM074,draw=white,line width=0.4pt] (6.28,0.00) rectangle (7.20,0.44);
\node[text=white] at (6.74,0.22) {100.1};
\draw[black,line width=0.95pt] (4.44,0) rectangle (5.36,3.52);
\node at (4.90,-0.40) {Decoder layers used};
\node[font=\sffamily\bfseries\fontsize{10}{12}\selectfont] at (10.30,4.08) {(b) \fohc{}};
\node[] at (8.46,3.75) {16};
\node[] at (9.38,3.75) {20};
\node[font=\sffamily\bfseries\fontsize{8.5}{10}\selectfont] at (10.30,3.75) {24};
\node[] at (11.22,3.75) {28};
\node[] at (12.14,3.75) {32};
\definecolor{fundmHM100}{rgb}{0.88771,0.93247,0.96827}
\filldraw[fill=fundmHM100,draw=white,line width=0.4pt] (8.00,3.08) rectangle (8.92,3.52);
\node[text=black] at (8.46,3.30) {55.4};
\definecolor{fundmHM101}{rgb}{0.37458,0.57648,0.73838}
\filldraw[fill=fundmHM101,draw=white,line width=0.4pt] (8.92,3.08) rectangle (9.84,3.52);
\node[text=black] at (9.38,3.30) {88.2};
\definecolor{fundmHM102}{rgb}{0.21853,0.41831,0.58823}
\filldraw[fill=fundmHM102,draw=white,line width=0.4pt] (9.84,3.08) rectangle (10.76,3.52);
\node[text=white] at (10.30,3.30) {98.2};
\definecolor{fundmHM103}{rgb}{0.22498,0.42573,0.59597}
\filldraw[fill=fundmHM103,draw=white,line width=0.4pt] (10.76,3.08) rectangle (11.68,3.52);
\node[text=white] at (11.22,3.30) {97.8};
\definecolor{fundmHM104}{rgb}{0.20885,0.40718,0.57662}
\filldraw[fill=fundmHM104,draw=white,line width=0.4pt] (11.68,3.08) rectangle (12.60,3.52);
\node[text=white] at (12.14,3.30) {98.9};
\definecolor{fundmHM110}{rgb}{0.62011,0.75768,0.86070}
\filldraw[fill=fundmHM110,draw=white,line width=0.4pt] (8.00,2.64) rectangle (8.92,3.08);
\node[text=black] at (8.46,2.86) {74.5};
\definecolor{fundmHM111}{rgb}{0.29162,0.50236,0.67594}
\filldraw[fill=fundmHM111,draw=white,line width=0.4pt] (8.92,2.64) rectangle (9.84,3.08);
\node[text=white] at (9.38,2.86) {93.2};
\definecolor{fundmHM112}{rgb}{0.23599,0.43839,0.60919}
\filldraw[fill=fundmHM112,draw=white,line width=0.4pt] (9.84,2.64) rectangle (10.76,3.08);
\node[text=white] at (10.30,2.86) {97.0};
\definecolor{fundmHM113}{rgb}{0.19479,0.39101,0.55975}
\filldraw[fill=fundmHM113,draw=white,line width=0.4pt] (10.76,2.64) rectangle (11.68,3.08);
\node[text=white] at (11.22,2.86) {99.9};
\definecolor{fundmHM114}{rgb}{0.19891,0.39574,0.56469}
\filldraw[fill=fundmHM114,draw=white,line width=0.4pt] (11.68,2.64) rectangle (12.60,3.08);
\node[text=white] at (12.14,2.86) {99.6};
\definecolor{fundmHM120}{rgb}{0.62354,0.75992,0.86212}
\filldraw[fill=fundmHM120,draw=white,line width=0.4pt] (8.00,2.20) rectangle (8.92,2.64);
\node[text=black] at (8.46,2.42) {74.3};
\definecolor{fundmHM121}{rgb}{0.28449,0.49416,0.66738}
\filldraw[fill=fundmHM121,draw=white,line width=0.4pt] (8.92,2.20) rectangle (9.84,2.64);
\node[text=white] at (9.38,2.42) {93.7};
\definecolor{fundmHM122}{rgb}{0.23011,0.43163,0.60213}
\filldraw[fill=fundmHM122,draw=white,line width=0.4pt] (9.84,2.20) rectangle (10.76,2.64);
\node[text=white] at (10.30,2.42) {97.4};
\definecolor{fundmHM123}{rgb}{0.18593,0.38082,0.54912}
\filldraw[fill=fundmHM123,draw=white,line width=0.4pt] (10.76,2.20) rectangle (11.68,2.64);
\node[text=white] at (11.22,2.42) {100.5};
\definecolor{fundmHM124}{rgb}{0.19613,0.39254,0.56135}
\filldraw[fill=fundmHM124,draw=white,line width=0.4pt] (11.68,2.20) rectangle (12.60,2.64);
\node[text=white] at (12.14,2.42) {99.8};
\definecolor{fundmHM130}{rgb}{0.36752,0.57112,0.73471}
\filldraw[fill=fundmHM130,draw=white,line width=0.4pt] (8.00,1.76) rectangle (8.92,2.20);
\node[text=black] at (8.46,1.98) {88.6};
\definecolor{fundmHM131}{rgb}{0.20926,0.40764,0.57711}
\filldraw[fill=fundmHM131,draw=white,line width=0.4pt] (8.92,1.76) rectangle (9.84,2.20);
\node[text=white] at (9.38,1.98) {98.9};
\definecolor{fundmHM132}{rgb}{0.18538,0.38019,0.54846}
\filldraw[fill=fundmHM132,draw=white,line width=0.4pt] (9.84,1.76) rectangle (10.76,2.20);
\node[text=white] at (10.30,1.98) {100.5};
\definecolor{fundmHM133}{rgb}{0.17803,0.37174,0.53964}
\filldraw[fill=fundmHM133,draw=white,line width=0.4pt] (10.76,1.76) rectangle (11.68,2.20);
\node[text=white] at (11.22,1.98) {101.0};
\definecolor{fundmHM134}{rgb}{0.18905,0.38441,0.55287}
\filldraw[fill=fundmHM134,draw=white,line width=0.4pt] (11.68,1.76) rectangle (12.60,2.20);
\node[text=white] at (12.14,1.98) {100.3};
\definecolor{fundmHM140}{rgb}{0.45684,0.63900,0.78116}
\filldraw[fill=fundmHM140,draw=white,line width=0.4pt] (8.00,1.32) rectangle (8.92,1.76);
\node[text=black] at (8.46,1.54) {83.7};
\definecolor{fundmHM141}{rgb}{0.24212,0.44544,0.61654}
\filldraw[fill=fundmHM141,draw=white,line width=0.4pt] (8.92,1.32) rectangle (9.84,1.76);
\node[text=white] at (9.38,1.54) {96.6};
\definecolor{fundmHM142}{rgb}{0.18932,0.38472,0.55319}
\filldraw[fill=fundmHM142,draw=white,line width=0.4pt] (9.84,1.32) rectangle (10.76,1.76);
\node[text=white] at (10.30,1.54) {100.2};
\definecolor{fundmHM143}{rgb}{0.18762,0.38276,0.55114}
\filldraw[fill=fundmHM143,draw=white,line width=0.4pt] (10.76,1.32) rectangle (11.68,1.76);
\node[text=white] at (11.22,1.54) {100.4};
\definecolor{fundmHM144}{rgb}{0.18932,0.38472,0.55319}
\filldraw[fill=fundmHM144,draw=white,line width=0.4pt] (11.68,1.32) rectangle (12.60,1.76);
\node[text=white] at (12.14,1.54) {100.2};
\definecolor{fundmHM150}{rgb}{0.61030,0.75128,0.85665}
\filldraw[fill=fundmHM150,draw=white,line width=0.4pt] (8.00,0.88) rectangle (8.92,1.32);
\node[text=black] at (8.46,1.10) {75.1};
\definecolor{fundmHM151}{rgb}{0.27187,0.47966,0.65225}
\filldraw[fill=fundmHM151,draw=white,line width=0.4pt] (8.92,0.88) rectangle (9.84,1.32);
\node[text=white] at (9.38,1.10) {94.6};
\definecolor{fundmHM152}{rgb}{0.16986,0.36234,0.52984}
\filldraw[fill=fundmHM152,draw=white,line width=0.4pt] (9.84,0.88) rectangle (10.76,1.32);
\node[text=white] at (10.30,1.10) {101.6};
\definecolor{fundmHM153}{rgb}{0.18745,0.38257,0.55094}
\filldraw[fill=fundmHM153,draw=white,line width=0.4pt] (10.76,0.88) rectangle (11.68,1.32);
\node[text=white] at (11.22,1.10) {100.4};
\definecolor{fundmHM154}{rgb}{0.18393,0.37852,0.54672}
\filldraw[fill=fundmHM154,draw=white,line width=0.4pt] (11.68,0.88) rectangle (12.60,1.32);
\node[text=white] at (12.14,1.10) {100.6};
\definecolor{fundmHM160}{rgb}{0.60461,0.74757,0.85429}
\filldraw[fill=fundmHM160,draw=white,line width=0.4pt] (8.00,0.44) rectangle (8.92,0.88);
\node[text=black] at (8.46,0.66) {75.4};
\definecolor{fundmHM161}{rgb}{0.32287,0.53718,0.71149}
\filldraw[fill=fundmHM161,draw=white,line width=0.4pt] (8.92,0.44) rectangle (9.84,0.88);
\node[text=black] at (9.38,0.66) {91.1};
\definecolor{fundmHM162}{rgb}{0.22617,0.42709,0.59740}
\filldraw[fill=fundmHM162,draw=white,line width=0.4pt] (9.84,0.44) rectangle (10.76,0.88);
\node[text=white] at (10.30,0.66) {97.7};
\definecolor{fundmHM163}{rgb}{0.20736,0.40546,0.57483}
\filldraw[fill=fundmHM163,draw=white,line width=0.4pt] (10.76,0.44) rectangle (11.68,0.88);
\node[text=white] at (11.22,0.66) {99.0};
\definecolor{fundmHM164}{rgb}{0.21154,0.41027,0.57984}
\filldraw[fill=fundmHM164,draw=white,line width=0.4pt] (11.68,0.44) rectangle (12.60,0.88);
\node[text=white] at (12.14,0.66) {98.7};
\definecolor{fundmHM170}{rgb}{0.76275,0.85071,0.91961}
\filldraw[fill=fundmHM170,draw=white,line width=0.4pt] (8.00,0.00) rectangle (8.92,0.44);
\node[text=black] at (8.46,0.22) {66.0};
\definecolor{fundmHM171}{rgb}{0.22784,0.42901,0.59940}
\filldraw[fill=fundmHM171,draw=white,line width=0.4pt] (8.92,0.00) rectangle (9.84,0.44);
\node[text=white] at (9.38,0.22) {97.6};
\definecolor{fundmHM172}{rgb}{0.19440,0.39056,0.55928}
\filldraw[fill=fundmHM172,draw=white,line width=0.4pt] (9.84,0.00) rectangle (10.76,0.44);
\node[text=white] at (10.30,0.22) {99.9};
\definecolor{fundmHM173}{rgb}{0.18938,0.38479,0.55326}
\filldraw[fill=fundmHM173,draw=white,line width=0.4pt] (10.76,0.00) rectangle (11.68,0.44);
\node[text=white] at (11.22,0.22) {100.2};
\definecolor{fundmHM174}{rgb}{0.20109,0.39825,0.56730}
\filldraw[fill=fundmHM174,draw=white,line width=0.4pt] (11.68,0.00) rectangle (12.60,0.44);
\node[text=white] at (12.14,0.22) {99.4};
\draw[black,line width=0.95pt] (9.84,0) rectangle (10.76,3.52);
\node at (10.30,-0.40) {Decoder layers used};
\definecolor{fundmCB0}{rgb}{0.93973,0.96655,0.98800}
\fill[fundmCB0] (13.28,0.0000) rectangle (13.53,0.0660);
\definecolor{fundmCB1}{rgb}{0.92918,0.95964,0.98400}
\fill[fundmCB1] (13.28,0.0640) rectangle (13.53,0.1300);
\definecolor{fundmCB2}{rgb}{0.91864,0.95273,0.98000}
\fill[fundmCB2] (13.28,0.1280) rectangle (13.53,0.1940);
\definecolor{fundmCB3}{rgb}{0.90809,0.94582,0.97600}
\fill[fundmCB3] (13.28,0.1920) rectangle (13.53,0.2580);
\definecolor{fundmCB4}{rgb}{0.89755,0.93891,0.97200}
\fill[fundmCB4] (13.28,0.2560) rectangle (13.53,0.3220);
\definecolor{fundmCB5}{rgb}{0.88700,0.93200,0.96800}
\fill[fundmCB5] (13.28,0.3200) rectangle (13.53,0.3860);
\definecolor{fundmCB6}{rgb}{0.87645,0.92509,0.96400}
\fill[fundmCB6] (13.28,0.3840) rectangle (13.53,0.4500);
\definecolor{fundmCB7}{rgb}{0.86591,0.91818,0.96000}
\fill[fundmCB7] (13.28,0.4480) rectangle (13.53,0.5140);
\definecolor{fundmCB8}{rgb}{0.85536,0.91127,0.95600}
\fill[fundmCB8] (13.28,0.5120) rectangle (13.53,0.5780);
\definecolor{fundmCB9}{rgb}{0.84482,0.90436,0.95200}
\fill[fundmCB9] (13.28,0.5760) rectangle (13.53,0.6420);
\definecolor{fundmCB10}{rgb}{0.83427,0.89745,0.94800}
\fill[fundmCB10] (13.28,0.6400) rectangle (13.53,0.7060);
\definecolor{fundmCB11}{rgb}{0.82373,0.89055,0.94400}
\fill[fundmCB11] (13.28,0.7040) rectangle (13.53,0.7700);
\definecolor{fundmCB12}{rgb}{0.81318,0.88364,0.94000}
\fill[fundmCB12] (13.28,0.7680) rectangle (13.53,0.8340);
\definecolor{fundmCB13}{rgb}{0.80264,0.87673,0.93600}
\fill[fundmCB13] (13.28,0.8320) rectangle (13.53,0.8980);
\definecolor{fundmCB14}{rgb}{0.78745,0.86682,0.92982}
\fill[fundmCB14] (13.28,0.8960) rectangle (13.53,0.9620);
\definecolor{fundmCB15}{rgb}{0.77073,0.85591,0.92291}
\fill[fundmCB15] (13.28,0.9600) rectangle (13.53,1.0260);
\definecolor{fundmCB16}{rgb}{0.75400,0.84500,0.91600}
\fill[fundmCB16] (13.28,1.0240) rectangle (13.53,1.0900);
\definecolor{fundmCB17}{rgb}{0.73727,0.83409,0.90909}
\fill[fundmCB17] (13.28,1.0880) rectangle (13.53,1.1540);
\definecolor{fundmCB18}{rgb}{0.72055,0.82318,0.90218}
\fill[fundmCB18] (13.28,1.1520) rectangle (13.53,1.2180);
\definecolor{fundmCB19}{rgb}{0.70382,0.81227,0.89527}
\fill[fundmCB19] (13.28,1.2160) rectangle (13.53,1.2820);
\definecolor{fundmCB20}{rgb}{0.68709,0.80136,0.88836}
\fill[fundmCB20] (13.28,1.2800) rectangle (13.53,1.3460);
\definecolor{fundmCB21}{rgb}{0.67036,0.79045,0.88145}
\fill[fundmCB21] (13.28,1.3440) rectangle (13.53,1.4100);
\definecolor{fundmCB22}{rgb}{0.65364,0.77955,0.87455}
\fill[fundmCB22] (13.28,1.4080) rectangle (13.53,1.4740);
\definecolor{fundmCB23}{rgb}{0.63691,0.76864,0.86764}
\fill[fundmCB23] (13.28,1.4720) rectangle (13.53,1.5380);
\definecolor{fundmCB24}{rgb}{0.62018,0.75773,0.86073}
\fill[fundmCB24] (13.28,1.5360) rectangle (13.53,1.6020);
\definecolor{fundmCB25}{rgb}{0.60345,0.74682,0.85382}
\fill[fundmCB25] (13.28,1.6000) rectangle (13.53,1.6660);
\definecolor{fundmCB26}{rgb}{0.58673,0.73591,0.84691}
\fill[fundmCB26] (13.28,1.6640) rectangle (13.53,1.7300);
\definecolor{fundmCB27}{rgb}{0.57000,0.72500,0.84000}
\fill[fundmCB27] (13.28,1.7280) rectangle (13.53,1.7940);
\definecolor{fundmCB28}{rgb}{0.55182,0.71118,0.83055}
\fill[fundmCB28] (13.28,1.7920) rectangle (13.53,1.8580);
\definecolor{fundmCB29}{rgb}{0.53364,0.69736,0.82109}
\fill[fundmCB29] (13.28,1.8560) rectangle (13.53,1.9220);
\definecolor{fundmCB30}{rgb}{0.51545,0.68355,0.81164}
\fill[fundmCB30] (13.28,1.9200) rectangle (13.53,1.9860);
\definecolor{fundmCB31}{rgb}{0.49727,0.66973,0.80218}
\fill[fundmCB31] (13.28,1.9840) rectangle (13.53,2.0500);
\definecolor{fundmCB32}{rgb}{0.47909,0.65591,0.79273}
\fill[fundmCB32] (13.28,2.0480) rectangle (13.53,2.1140);
\definecolor{fundmCB33}{rgb}{0.46091,0.64209,0.78327}
\fill[fundmCB33] (13.28,2.1120) rectangle (13.53,2.1780);
\definecolor{fundmCB34}{rgb}{0.44273,0.62827,0.77382}
\fill[fundmCB34] (13.28,2.1760) rectangle (13.53,2.2420);
\definecolor{fundmCB35}{rgb}{0.42455,0.61445,0.76436}
\fill[fundmCB35] (13.28,2.2400) rectangle (13.53,2.3060);
\definecolor{fundmCB36}{rgb}{0.40636,0.60064,0.75491}
\fill[fundmCB36] (13.28,2.3040) rectangle (13.53,2.3700);
\definecolor{fundmCB37}{rgb}{0.38818,0.58682,0.74545}
\fill[fundmCB37] (13.28,2.3680) rectangle (13.53,2.4340);
\definecolor{fundmCB38}{rgb}{0.37000,0.57300,0.73600}
\fill[fundmCB38] (13.28,2.4320) rectangle (13.53,2.4980);
\definecolor{fundmCB39}{rgb}{0.35182,0.55918,0.72655}
\fill[fundmCB39] (13.28,2.4960) rectangle (13.53,2.5620);
\definecolor{fundmCB40}{rgb}{0.33364,0.54536,0.71709}
\fill[fundmCB40] (13.28,2.5600) rectangle (13.53,2.6260);
\definecolor{fundmCB41}{rgb}{0.31636,0.53082,0.70564}
\fill[fundmCB41] (13.28,2.6240) rectangle (13.53,2.6900);
\definecolor{fundmCB42}{rgb}{0.30182,0.51409,0.68818}
\fill[fundmCB42] (13.28,2.6880) rectangle (13.53,2.7540);
\definecolor{fundmCB43}{rgb}{0.28727,0.49736,0.67073}
\fill[fundmCB43] (13.28,2.7520) rectangle (13.53,2.8180);
\definecolor{fundmCB44}{rgb}{0.27273,0.48064,0.65327}
\fill[fundmCB44] (13.28,2.8160) rectangle (13.53,2.8820);
\definecolor{fundmCB45}{rgb}{0.25818,0.46391,0.63582}
\fill[fundmCB45] (13.28,2.8800) rectangle (13.53,2.9460);
\definecolor{fundmCB46}{rgb}{0.24364,0.44718,0.61836}
\fill[fundmCB46] (13.28,2.9440) rectangle (13.53,3.0100);
\definecolor{fundmCB47}{rgb}{0.22909,0.43045,0.60091}
\fill[fundmCB47] (13.28,3.0080) rectangle (13.53,3.0740);
\definecolor{fundmCB48}{rgb}{0.21455,0.41373,0.58345}
\fill[fundmCB48] (13.28,3.0720) rectangle (13.53,3.1380);
\definecolor{fundmCB49}{rgb}{0.20000,0.39700,0.56600}
\fill[fundmCB49] (13.28,3.1360) rectangle (13.53,3.2020);
\definecolor{fundmCB50}{rgb}{0.18545,0.38027,0.54855}
\fill[fundmCB50] (13.28,3.2000) rectangle (13.53,3.2660);
\definecolor{fundmCB51}{rgb}{0.17091,0.36355,0.53109}
\fill[fundmCB51] (13.28,3.2640) rectangle (13.53,3.3300);
\definecolor{fundmCB52}{rgb}{0.15636,0.34682,0.51364}
\fill[fundmCB52] (13.28,3.3280) rectangle (13.53,3.3940);
\definecolor{fundmCB53}{rgb}{0.14182,0.33009,0.49618}
\fill[fundmCB53] (13.28,3.3920) rectangle (13.53,3.4580);
\definecolor{fundmCB54}{rgb}{0.12727,0.31336,0.47873}
\fill[fundmCB54] (13.28,3.4560) rectangle (13.53,3.5220);
\draw[black!70,line width=0.4pt] (13.53,0.0000) -- (13.62,0.0000);
\node[anchor=west,font=\sffamily\fontsize{8}{9}\selectfont] at (13.69,0.0000) {50};
\draw[black!70,line width=0.4pt] (13.53,0.6400) -- (13.62,0.6400);
\node[anchor=west,font=\sffamily\fontsize{8}{9}\selectfont] at (13.69,0.6400) {60};
\draw[black!70,line width=0.4pt] (13.53,1.2800) -- (13.62,1.2800);
\node[anchor=west,font=\sffamily\fontsize{8}{9}\selectfont] at (13.69,1.2800) {70};
\draw[black!70,line width=0.4pt] (13.53,1.9200) -- (13.62,1.9200);
\node[anchor=west,font=\sffamily\fontsize{8}{9}\selectfont] at (13.69,1.9200) {80};
\draw[black!70,line width=0.4pt] (13.53,2.5600) -- (13.62,2.5600);
\node[anchor=west,font=\sffamily\fontsize{8}{9}\selectfont] at (13.69,2.5600) {90};
\draw[black!70,line width=0.4pt] (13.53,3.2000) -- (13.62,3.2000);
\node[anchor=west,font=\sffamily\fontsize{8}{9}\selectfont] at (13.69,3.2000) {100};
\draw[black!70,line width=0.4pt] (13.53,3.5200) -- (13.62,3.5200);
\node[anchor=west,font=\sffamily\fontsize{8}{9}\selectfont] at (13.69,3.5200) {105};
\node[rotate=90] at (14.70,1.76) {Teacher accuracy retained (\%)};
\end{tikzpicture}%
}
\caption{Teacher-relative accuracy by benchmark and depth (seed 0):
$100\times$ student accuracy divided by teacher accuracy, computed
from three-decimal scores. Both panels share the same color scale,
without clipping values above 100\%. Black outlines mark fixed
layer 24. Per-benchmark scores are in the reproduction data.}
\label{fig:benchmark-depth}
\end{figure*}

\paragraph{Per-Exit Depth Benchmark Performance.}
At layer 20, MMLU-Pro retains only 86--88\% of teacher accuracy
versus 94--95\% for the eight-benchmark mean. Figure~\ref{fig:benchmark-depth}
shows this variation across benchmarks and depths.

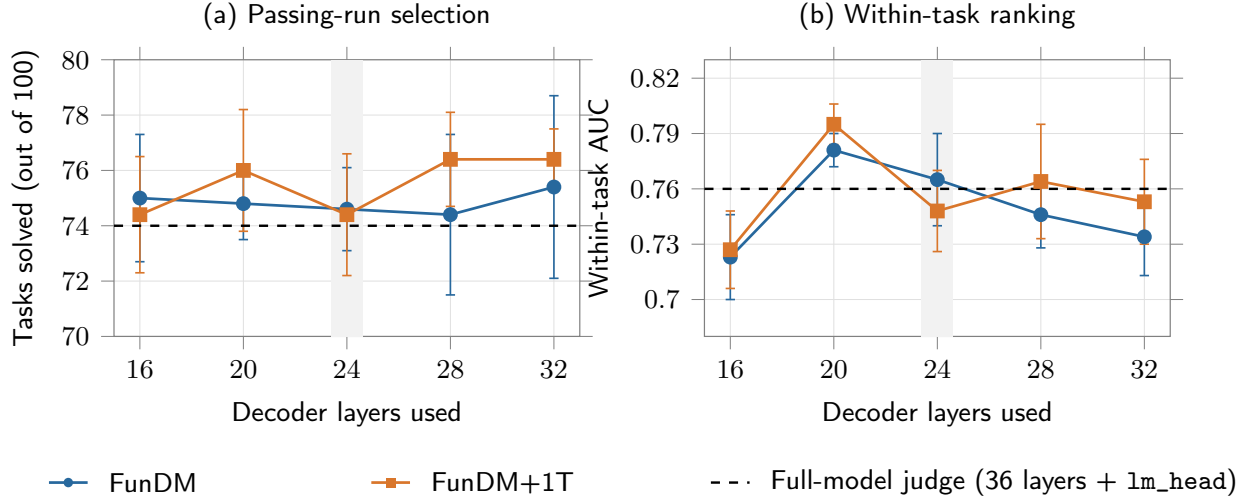
\begin{figure*}[!t]
\centering
\resizebox{\linewidth}{!}{%
\begin{tikzpicture}
\begin{groupplot}[group style={group size=2 by 1,horizontal sep=1.45cm},width=7.0cm,height=4.8cm,
/tikz/font=\sffamily\footnotesize,xmin=15,xmax=33,xtick={16,20,24,28,32},xlabel={Decoder layers used},
tick align=outside,axis line style={black!50},tick style={black!50},grid=major,grid style={black!12,line width=0.2pt}]
\nextgroupplot[ylabel={Tasks solved (out of 100)},title={(a) Passing-run selection},ymin=70,ymax=80,ytick={70,72,74,76,78,80}]
\path[fill=black!5] (axis cs:23.4,70) rectangle (axis cs:24.6,80);
\draw[fundmBlue,line width=0.55pt] (axis cs:16,72.700) -- (axis cs:16,77.300);
\draw[fundmBlue,line width=0.55pt] (axis cs:15.82,72.700) -- (axis cs:16.18,72.700);
\draw[fundmBlue,line width=0.55pt] (axis cs:15.82,77.300) -- (axis cs:16.18,77.300);
\draw[fundmBlue,line width=0.55pt] (axis cs:20,73.500) -- (axis cs:20,76.100);
\draw[fundmBlue,line width=0.55pt] (axis cs:19.82,73.500) -- (axis cs:20.18,73.500);
\draw[fundmBlue,line width=0.55pt] (axis cs:19.82,76.100) -- (axis cs:20.18,76.100);
\draw[fundmBlue,line width=0.55pt] (axis cs:24,73.100) -- (axis cs:24,76.100);
\draw[fundmBlue,line width=0.55pt] (axis cs:23.82,73.100) -- (axis cs:24.18,73.100);
\draw[fundmBlue,line width=0.55pt] (axis cs:23.82,76.100) -- (axis cs:24.18,76.100);
\draw[fundmBlue,line width=0.55pt] (axis cs:28,71.500) -- (axis cs:28,77.300);
\draw[fundmBlue,line width=0.55pt] (axis cs:27.82,71.500) -- (axis cs:28.18,71.500);
\draw[fundmBlue,line width=0.55pt] (axis cs:27.82,77.300) -- (axis cs:28.18,77.300);
\draw[fundmBlue,line width=0.55pt] (axis cs:32,72.100) -- (axis cs:32,78.700);
\draw[fundmBlue,line width=0.55pt] (axis cs:31.82,72.100) -- (axis cs:32.18,72.100);
\draw[fundmBlue,line width=0.55pt] (axis cs:31.82,78.700) -- (axis cs:32.18,78.700);
\addplot+[color=fundmBlue,mark=*,mark size=2pt,line width=0.9pt,mark options={solid,fill=fundmBlue}] coordinates {
(16,75.000) 
(20,74.800) 
(24,74.600) 
(28,74.400) 
(32,75.400) 
};
\draw[fundmOrange,line width=0.55pt] (axis cs:16,72.300) -- (axis cs:16,76.500);
\draw[fundmOrange,line width=0.55pt] (axis cs:15.82,72.300) -- (axis cs:16.18,72.300);
\draw[fundmOrange,line width=0.55pt] (axis cs:15.82,76.500) -- (axis cs:16.18,76.500);
\draw[fundmOrange,line width=0.55pt] (axis cs:20,73.800) -- (axis cs:20,78.200);
\draw[fundmOrange,line width=0.55pt] (axis cs:19.82,73.800) -- (axis cs:20.18,73.800);
\draw[fundmOrange,line width=0.55pt] (axis cs:19.82,78.200) -- (axis cs:20.18,78.200);
\draw[fundmOrange,line width=0.55pt] (axis cs:24,72.200) -- (axis cs:24,76.600);
\draw[fundmOrange,line width=0.55pt] (axis cs:23.82,72.200) -- (axis cs:24.18,72.200);
\draw[fundmOrange,line width=0.55pt] (axis cs:23.82,76.600) -- (axis cs:24.18,76.600);
\draw[fundmOrange,line width=0.55pt] (axis cs:28,74.700) -- (axis cs:28,78.100);
\draw[fundmOrange,line width=0.55pt] (axis cs:27.82,74.700) -- (axis cs:28.18,74.700);
\draw[fundmOrange,line width=0.55pt] (axis cs:27.82,78.100) -- (axis cs:28.18,78.100);
\draw[fundmOrange,line width=0.55pt] (axis cs:32,75.300) -- (axis cs:32,77.500);
\draw[fundmOrange,line width=0.55pt] (axis cs:31.82,75.300) -- (axis cs:32.18,75.300);
\draw[fundmOrange,line width=0.55pt] (axis cs:31.82,77.500) -- (axis cs:32.18,77.500);
\addplot+[color=fundmOrange,mark=square*,mark size=2pt,line width=0.9pt,mark options={solid,fill=fundmOrange}] coordinates {
(16,74.400) 
(20,76.000) 
(24,74.400) 
(28,76.400) 
(32,76.400) 
};
\addplot[black,dashed,line width=0.8pt,domain=15:33,samples=2] {74.0};
\nextgroupplot[ylabel={Within-task AUC},title={(b) Within-task ranking},ymin=0.68,ymax=0.83,ytick={0.70,0.73,0.76,0.79,0.82}]
\path[fill=black!5] (axis cs:23.4,0.68) rectangle (axis cs:24.6,0.83);
\draw[fundmBlue,line width=0.55pt] (axis cs:16,0.700) -- (axis cs:16,0.746);
\draw[fundmBlue,line width=0.55pt] (axis cs:15.82,0.700) -- (axis cs:16.18,0.700);
\draw[fundmBlue,line width=0.55pt] (axis cs:15.82,0.746) -- (axis cs:16.18,0.746);
\draw[fundmBlue,line width=0.55pt] (axis cs:20,0.772) -- (axis cs:20,0.790);
\draw[fundmBlue,line width=0.55pt] (axis cs:19.82,0.772) -- (axis cs:20.18,0.772);
\draw[fundmBlue,line width=0.55pt] (axis cs:19.82,0.790) -- (axis cs:20.18,0.790);
\draw[fundmBlue,line width=0.55pt] (axis cs:24,0.740) -- (axis cs:24,0.790);
\draw[fundmBlue,line width=0.55pt] (axis cs:23.82,0.740) -- (axis cs:24.18,0.740);
\draw[fundmBlue,line width=0.55pt] (axis cs:23.82,0.790) -- (axis cs:24.18,0.790);
\draw[fundmBlue,line width=0.55pt] (axis cs:28,0.728) -- (axis cs:28,0.764);
\draw[fundmBlue,line width=0.55pt] (axis cs:27.82,0.728) -- (axis cs:28.18,0.728);
\draw[fundmBlue,line width=0.55pt] (axis cs:27.82,0.764) -- (axis cs:28.18,0.764);
\draw[fundmBlue,line width=0.55pt] (axis cs:32,0.713) -- (axis cs:32,0.755);
\draw[fundmBlue,line width=0.55pt] (axis cs:31.82,0.713) -- (axis cs:32.18,0.713);
\draw[fundmBlue,line width=0.55pt] (axis cs:31.82,0.755) -- (axis cs:32.18,0.755);
\addplot+[color=fundmBlue,mark=*,mark size=2pt,line width=0.9pt,mark options={solid,fill=fundmBlue}] coordinates {
(16,0.723) 
(20,0.781) 
(24,0.765) 
(28,0.746) 
(32,0.734) 
};
\draw[fundmOrange,line width=0.55pt] (axis cs:16,0.706) -- (axis cs:16,0.748);
\draw[fundmOrange,line width=0.55pt] (axis cs:15.82,0.706) -- (axis cs:16.18,0.706);
\draw[fundmOrange,line width=0.55pt] (axis cs:15.82,0.748) -- (axis cs:16.18,0.748);
\draw[fundmOrange,line width=0.55pt] (axis cs:20,0.784) -- (axis cs:20,0.806);
\draw[fundmOrange,line width=0.55pt] (axis cs:19.82,0.784) -- (axis cs:20.18,0.784);
\draw[fundmOrange,line width=0.55pt] (axis cs:19.82,0.806) -- (axis cs:20.18,0.806);
\draw[fundmOrange,line width=0.55pt] (axis cs:24,0.726) -- (axis cs:24,0.770);
\draw[fundmOrange,line width=0.55pt] (axis cs:23.82,0.726) -- (axis cs:24.18,0.726);
\draw[fundmOrange,line width=0.55pt] (axis cs:23.82,0.770) -- (axis cs:24.18,0.770);
\draw[fundmOrange,line width=0.55pt] (axis cs:28,0.733) -- (axis cs:28,0.795);
\draw[fundmOrange,line width=0.55pt] (axis cs:27.82,0.733) -- (axis cs:28.18,0.733);
\draw[fundmOrange,line width=0.55pt] (axis cs:27.82,0.795) -- (axis cs:28.18,0.795);
\draw[fundmOrange,line width=0.55pt] (axis cs:32,0.730) -- (axis cs:32,0.776);
\draw[fundmOrange,line width=0.55pt] (axis cs:31.82,0.730) -- (axis cs:32.18,0.730);
\draw[fundmOrange,line width=0.55pt] (axis cs:31.82,0.776) -- (axis cs:32.18,0.776);
\addplot+[color=fundmOrange,mark=square*,mark size=2pt,line width=0.9pt,mark options={solid,fill=fundmOrange}] coordinates {
(16,0.727) 
(20,0.795) 
(24,0.748) 
(28,0.764) 
(32,0.753) 
};
\addplot[black,dashed,line width=0.8pt,domain=15:33,samples=2] {0.76};
\end{groupplot}
\begin{scope}[shift={($(group c1r1.south)!0.5!(group c2r1.south)+(-1.40cm,-1.70cm)$)},font=\sffamily\footnotesize]
\draw[fundmBlue,line width=0.9pt] (-5.50,0) -- (-5.05,0);
\fill[fundmBlue] (-5.275,0) circle (0.055cm);
\node[anchor=west] at (-4.94,0) {\foh{}};
\draw[fundmOrange,line width=0.9pt] (-1.75,0) -- (-1.30,0);
\fill[fundmOrange] (-1.58,-0.055) rectangle (-1.47,0.055);
\node[anchor=west] at (-1.19,0) {\fohc{}};
\draw[black,dashed,line width=0.8pt] (2.20,0) -- (2.65,0);
\node[anchor=west] at (2.76,0) {Full-model judge (36 layers + \texttt{lm\_head})};
\end{scope}
\end{tikzpicture}%
}
\caption{CoderForge exit-depth sweep: five-seed means $\pm$ sample
standard deviations, not task-sampling confidence intervals.
Dashed lines show the full-model judge (74 tasks; AUC 0.760).
Shading marks fixed layer 24.}
\label{fig:agent-depth}
\end{figure*}

\paragraph{Agent Selection.}
Solved-task means stay between 74.4 and 76.4 across layers 16--32
(Figure~\ref{fig:agent-depth}). AUC peaks at layer 20 and falls to
0.72--0.73 at layer 16: selecting one passing run can remain
effective despite a less accurate overall ranking. The main
configuration is fixed at layer 24; its paired uncertainty
analysis does not cover other exits.

\paragraph{Training Data and Exit Depth.}
Adding 84{,}757 rows from RACE, MathQA, Cosmos QA, QASC, and LogiQA
\cite{lai2017race,amini2019mathqa,huang2019cosmosqa,khot2019qasc,
liu2020logiqa} expands the pool to 123k. At layer 24, mean accuracy
rises by 0.2/0.1 percentage points for \foh{}/\fohc{}; at layer 20,
by 1.1/1.0. At layer 16, \foh{} loses about 1.4 points and
\fohc{} gains about 3.6. Neither 16- nor 20-layer heads match the
teacher. This single-seed comparison also changes the data sources and
validation splits, so it does not isolate the effect of adding more data (Table~\ref{tab:data}, Appendix~\ref{app:data-details}).

\section{Conclusion}
\foh{} reuses pretrained decision capability with a small head
trained on teacher probabilities, without correct-answer labels
for fitting or checkpoint selection. At 24 layers, \fohc{}
retains 99.2\% of the full model's mean MCQ accuracy; separately
trained agent heads show comparable run selection. Skipping
12 layers and the LM head reduces compute and the layers requiring
KV storage. On MMLU-Pro, the heads provide $1.56$--$1.57\times$
vLLM throughput, about 33\% lower peak GPU allocator memory,
and 38\% lower first-decision latency. Future work could test
other backbones and workloads.

\FloatBarrier
\section*{Limitations}
\label{sec:limits}

\foh{} recovers much of the full model's performance at several
fixed exit depths. We have tested only Qwen3-8B, but the provided
recipe can be adapted to other open-weights language models.
Future work could evaluate larger models, adaptive exits, and
more training seeds, and select alternative exit depths on
development data.

Our distillation ablation compares teacher probabilities with teacher
top-choice and correct-option targets, but not with Tuned Lens-style
objectives.

For agent selection, teacher probabilities are often close to
zero or one, and task-bootstrap confidence intervals are wide.

The separate DeepSWE comparison in Appendix~\ref{app:deepswe}
uses CLM's released embeddings without run text \cite{kwok2026clm}.
It tests \fohce{} and \fohcce{} with separate reading, but cannot
test joint encoding or teacher-based KD. Regenerating CLM's full
corpus exceeds our budget and would not reproduce its stochastic
runs exactly; Appendix~\ref{app:deepswe} gives the corpus details.

Efficiency is measured on one GPU. The memory and first-decision
results cover one multiple-choice workload and engine calls;
production queueing, HTTP overhead, and non-PyTorch GPU memory
are outside these measurements.

\section*{Ethical Considerations}
FunDM learns from the frozen teacher's predictions and may inherit its
errors and biases; benchmark accuracy does not establish reliability
for consequential decisions. In coding-agent selection, a highly
ranked run may still contain an incorrect or insecure change and
requires further testing and review before deployment. We measured
lower inference resource use for one workload but did not measure
total energy use or emissions, including teacher caching and training.

\section*{Acknowledgements}
The author thanks Bertrand Serlet, Greg Gilley, Insop Song, and Manny Ko
for their valuable reviews and advice.

\clearpage
\appendix
\section{Implementation and Training}
\label{app:impl}

\subsection{Artifact Release and Licenses}
The planned release contains our code, figures, aggregate results, and
task/run identifiers. It does not redistribute model weights, benchmark
questions, coding-agent trajectories, or embeddings; scripts obtain
these inputs from their original hosts. Qwen3-8B and CLM are released
under Apache License 2.0 by their creators \cite{yang2025qwen3,kwok2026clm}.
The CLM code is linked as an upstream submodule with its license and
attribution retained. Our original code is also released under Apache
License 2.0. The release repository is
\url{https://github.com/Ujjawal-K-Panchal/FunDM}.

\subsection{Prompt Formats and State Extraction}
\label{sec:prompt}
The teacher and both FunDM variants use the multi-line prompt below
because it gives the frozen model 0.744 accuracy on 1{,}903 held-out
validation rows (chance 0.194), versus 0.570 with the initial
single-line prompt. The difference is 17.4 percentage points
(95\% bootstrap interval 15.4 to 19.5).
\begin{center}
\footnotesize\ttfamily\frenchspacing
\begin{tabular}{p{\dimexpr\linewidth-2\tabcolsep\relax}}
The following is a multiple choice question.\\
Answer with the letter of the correct option.\\[3pt]
Question: <question>\\
Options:\\
A. <option 1>\\
\ldots\\
Answer:
\end{tabular}
\end{center}
The two instruction sentences are joined by a space. The initial
single-line teacher prompt was (wrapped here for display):
\begin{center}
\footnotesize\ttfamily\frenchspacing
\begin{tabular}{p{\dimexpr\linewidth-2\tabcolsep\relax}}
<question> answer options: (a) <option 1>\\
(b) <option 2> \ldots\\
:: answer in option character only (without brackets): (
\end{tabular}
\end{center}
Under this prompt, the teacher predicts \texttt{a}--\texttt{j}
after the final open parenthesis.
Option labels, instructions, and delimiters change together, so
this comparison tests the prompt recipe, not any one change.

For FunDM, \verb*| A|--\verb*| J| are single tokens. The teacher reads
the final normalized state at the last colon; option letters hold
51\% of token probability on MMLU-Pro and 77--96\% on other sets
before renormalization. We locate option ends by character offsets
(verified on 500 MMLU-Pro rows), store layer outputs in bfloat16,
and normalize student states by their root mean square.

\fohce{} and \fohcce{} were trained using final-layer normalized
states at the closing delimiters and final colon of this single-line prompt:
\begin{center}
\scriptsize
\texttt{<question> rubric: <option 1> ::}
\texttt{\ldots{} :: <option $K$> :: answer:}
\end{center}
We verify that the delimiter \verb| ::| tokenizes to a single token.
Recorded boundaries avoid mistaking delimiters inside
options for option ends. Padded options have logits of $-\infty$.
For MCQ evaluation and training, no FunDM row exceeds 4{,}096 tokens,
no CE row exceeds 2{,}048, and no rows are dropped.
BoolQ places its passage and question together, with ``no'' and ``yes'' options.

As a prompt sensitivity check, the teacher scores 45.2\% on MMLU-Pro with the original option
order and 45.1\% with shuffled options. This check does not test
whether student predictions are robust to option order.

\subsection{Architecture and Optimization}
\fohc{} uses 512-dimensional states, option/answer type embeddings,
and one pre-norm, 8-head bidirectional transformer layer with a
2{,}048-dimensional GELU feed-forward block and dropout 0.1.
A final layer norm precedes the 128-dimensional scoring projections;
no positional encoding is added. MCQ heads train for ten epochs in
batches of 256 with AdamW (weight decay 0.01) and a one-cycle schedule,
using the first 5\% of steps for warmup. Peak learning rates are
$10^{-3}$ for the base head and $5\times10^{-4}$ for +1T and
CLM-shaped CE heads. Validation NLL selects CE checkpoints;
validation KL selects FunDM checkpoints. Trainable parameter counts
are given in Section~\ref{sec:method}.

\subsection{Data Splits and Overlap}
For the expanded pool, we first remove 18 source questions sharing a 13-word span
with an evaluation question. Added rows are split by dataset ID;
664 validation rows share question text with training, including one
identical full prompt. Evaluation uses test splits for
MMLU-Pro, MMLU, and OpenBookQA, and validation splits for
CommonsenseQA, PIQA, HellaSwag, WinoGrande, and BoolQ. No evaluation
split is used for checkpoint selection.

Normalized exact matching finds one MMLU question in the original
training pool and none from the other evaluations; paraphrases and
backbone/CLM pretraining overlap are not checked. Dataset revisions,
source counts, and construction details will accompany the public
implementation.

\section{Additional Ablation Results}
\label{app:ablation-details}
\subsection{Original Supervised CE Recipe}
\label{app:ce-details}
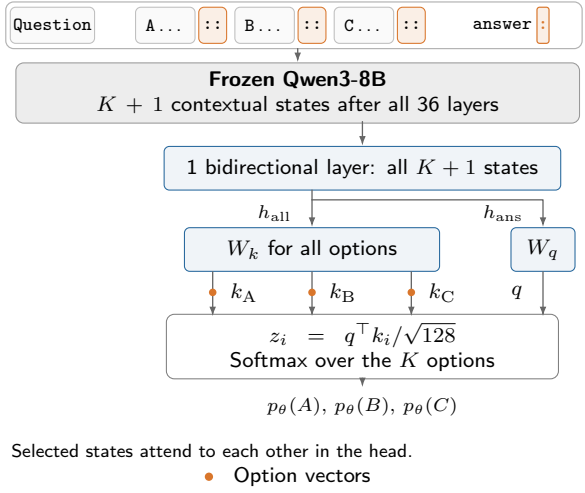
\begin{figure*}[!t]
\centering
\resizebox{\linewidth}{!}{%
\begin{tikzpicture}[
  x=1cm,y=1cm,
  font=\sffamily\fontsize{8}{9.5}\selectfont,
  line cap=round,line join=round,
  token/.style={draw=black!25,fill=black!2,rounded corners=2pt,
    minimum height=0.48cm,inner sep=1pt,align=center,
    font=\sffamily\fontsize{7}{8.5}\selectfont},
  frozen/.style={draw=black!40,fill=black!7,rounded corners=3pt,
    minimum height=0.76cm,inner sep=3pt,align=center},
  learned/.style={draw=fundmBlue,fill=fundmBlue!7,rounded corners=2pt,
    minimum height=0.58cm,inner sep=3pt,align=center},
  output/.style={draw=black!50,fill=white,rounded corners=3pt,
    minimum height=0.75cm,inner sep=3pt,align=center},
  flow/.style={-{Latex[length=1.3mm,width=0.9mm]},draw=black!60,line width=0.5pt},
  wire/.style={draw=black!60,line width=0.5pt},
  cue/.style={draw=black!25,fill=black!2,line width=0.45pt},
  skipped/.style={draw=black!35,text=black!55,dashed,rounded corners=3pt,
    fill=white,inner sep=3pt,align=center},
  heading/.style={anchor=west,font=\sffamily\fontsize{9.2}{11}\selectfont\bfseries},
  subtitle/.style={anchor=west,text=black!60,font=\sffamily\fontsize{7.5}{9}\selectfont},
  footer/.style={anchor=north west,align=left,font=\sffamily\fontsize{7.5}{9}\selectfont},
  probability/.style={font=\sffamily\fontsize{7.5}{9}\selectfont}
]
\path[use as bounding box] (0,-7.25) rectangle (16.46,0.20);
\draw[black!15,line width=0.4pt] (8.16,0.10) -- (8.16,-6.65);

\foreach \shift/\tag/\titletext in {
  0/ce/{(a) \fohce{}},
  8.35/ceone/{(b) \fohcce{}}}{
  \begin{scope}[xshift=\shift cm]
    \node[heading] at (0.04,0) {\titletext};
    \node[subtitle] at (0.04,-0.37) {One joint causal prompt};
    \draw[draw=black!30,rounded corners=3pt,line width=0.45pt]
      (0.10,-1.31) rectangle (7.76,-0.69);
    \node[token,text width=1.05cm] at (0.72,-1.00) {\texttt{Question}};
    \foreach \x/\sym/\delimiter in {2.22/A/2.85,3.54/B/4.17,4.86/C/5.49} {
      \node[token,text width=0.72cm] at (\x,-1.00) {\texttt{\sym}\,\ldots{}};
      \node[token,draw=fundmOrange,fill=fundmOrange!12,text width=0.30cm]
        at (\delimiter,-1.00) {\texttt{::}};
    }
    \node[font=\sffamily\fontsize{7}{8.5}\selectfont] at (6.70,-1.00) {\texttt{answer}};
    \draw[cue,draw=fundmOrange,fill=fundmOrange!12,rounded corners=1pt] (7.16,-1.24) rectangle (7.32,-0.76);
    \node[text=fundmOrange,font=\sffamily\fontsize{7}{8.5}\selectfont\ttfamily]
      at (7.24,-1.00) {:};
    \node[frozen,text width=7.28cm] (\tag-model) at (3.98,-1.90)
      {\textbf{Frozen Qwen3-8B}\\$K+1$ contextual states after all 36 layers};
    \draw[flow] (3.98,-1.31) -- (\tag-model.north);
    \node[learned,text width=3.18cm] (\tag-key) at (4.17,-3.98)
      {$W_k$ for all options};
    \node[learned,text width=0.64cm] (\tag-query) at (7.24,-3.98) {$W_q$};
    \node[output,text width=5.00cm] (\tag-score) at (4.84,-5.27)
      {$z_i=q^\top k_i/\sqrt{128}$\\Softmax over the $K$ options};
    \foreach \x/\sym in {2.85/A,4.17/B,5.49/C} {
      \draw[flow] (\tag-key.south -| \x,0) -- (\x,-4.87);
      \fill[fundmOrange] (\x,-4.55) circle (0.05);
      \node[anchor=west] at (\x+0.10,-4.55) {$k_{\mathrm{\sym}}$};
    }
    \draw[flow] (\tag-query.south) -- (7.24,-4.87);
    \node[anchor=east] at (7.10,-4.55) {$q$};
    \node[probability] (\tag-probs) at (4.84,-6.07)
      {$p_\theta(A),\,p_\theta(B),\,p_\theta(C)$};
    \draw[flow] (\tag-score.south) -- (\tag-probs.north);
  \end{scope}
}

\draw[wire] (4.17,-2.28) -- (4.17,-2.57);
\draw[flow] (4.17,-2.57) -- (ce-key.north);
\node[probability,anchor=east] at (4.03,-3.13) {$h_{\mathrm{all}}$};
\draw[flow] (4.17,-2.57) -- (7.24,-2.57) -- (ce-query.north);
\node[probability,anchor=east] at (7.11,-3.13) {$h_{\mathrm{ans}}$};
\node[subtitle] at (0.20,-3.52) {Direct projections};
\node[footer,text width=7.75cm] at (0.04,-6.43)
  {$h_B$ has read the question, A and B, not C.};

\begin{scope}[xshift=8.35cm]
  \node[learned,text width=5.06cm] (ce-mixing) at (4.84,-2.90)
    {1 bidirectional layer: all $K+1$ states};
  \draw[flow] (4.84,-2.28) -- (ce-mixing.north);
  \draw[wire] (4.17,-3.19) -- (4.17,-3.32);
  \draw[flow] (4.17,-3.32) -- (ceone-key.north);
  \node[probability,anchor=east] at (4.03,-3.50) {$h_{\mathrm{all}}$};
  \draw[flow] (4.17,-3.32) -- (7.24,-3.32) -- (ceone-query.north);
  \node[probability,anchor=east] at (7.11,-3.50) {$h_{\mathrm{ans}}$};
  \node[footer,text width=7.75cm] at (0.04,-6.43)
    {Selected states attend to each other in the head.};
\end{scope}

\node[frozen,text width=0.48cm,minimum height=0.28cm,inner sep=0pt] at (0.35,-7.00) {};
\node[anchor=west] at (0.71,-7.00) {Frozen model};
\node[learned,text width=0.48cm,minimum height=0.28cm,inner sep=0pt] at (5.75,-7.00) {};
\node[anchor=west] at (6.11,-7.00) {Trained head};
\fill[fundmOrange] (11.15,-7.00) circle (0.055);
\node[anchor=west] at (11.34,-7.00) {Option vectors};
\end{tikzpicture}
}
\caption{These joint-encoding heads use correct-option labels,
categorical cross-entropy, and final-layer contextual states.
\textbf{(a)} \fohce{} projects readouts directly.
\textbf{(b)} \fohcce{} first mixes all $K+1$ readouts bidirectionally.
$h_{\mathrm{all}}$ stacks option states for the shared $W_k$;
the answer state feeds $W_q$.}
\label{fig:encoding}

\end{figure*}

The original \fohce{} baseline introduced in Section~\ref{sec:setup}
uses all 36 layers, final-layer normalized
states, and the single-line prompt in Appendix~\ref{sec:prompt},
with 7.57B total and 1.05M trainable parameters.
Table~\ref{tab:ce-sets} reports accuracy and Figure~\ref{fig:encoding}
shows the joint CE heads; timing is in Section~\ref{sec:efficiency}
and Appendix~\ref{app:timing}. The full-model reference uses the
FunDM prompt. Because prompt, readout, normalization, and depth
differ, Table~\ref{tab:target} provides the matched target comparison.

\begin{table*}[!t]
\centering\small
\begin{tabular}{lcc}
\hline
Benchmark & \fohce{}, original layer-36 recipe & Full-model reference \\
\hline
MMLU-Pro & 0.304 & 0.451 \\
MMLU & 0.543 & 0.706 \\
OpenBookQA & 0.580 & 0.856 \\
CommonsenseQA & 0.616 & 0.792 \\
PIQA & 0.643 & 0.854 \\
HellaSwag & 0.490 & 0.827 \\
WinoGrande & 0.506 & 0.696 \\
BoolQ & 0.685 & 0.870 \\
\hline
Mean & 0.546 & 0.757 \\
\hline
\end{tabular}
\caption{Eight-benchmark accuracy of the original supervised CE
recipe (seed 0). The full-model reference uses the \foh{} prompt.
This is not a matched training-target comparison; Table~\ref{tab:target}
provides that ablation.}
\label{tab:ce-sets}
\end{table*}

On CoderForge, action-trained \fohce{} solves $61.0\pm1.6$ tasks
with AUC $0.554\pm0.009$; \fohcce{} solves $57.8\pm3.6$ with AUC
$0.525\pm0.005$ over five seeds. These heads learn from passing
actions, whereas FunDM learns whole-run judge probabilities;
inputs, targets, and run scoring differ.

\subsection{Head Architecture Diagnostics}
\label{app:architecture-details}
Table~\ref{tab:gold-heads} compares heads on identical joint
final-layer states. Lower NLL gives more probability to correct
answers; lower ECE means confidence better matches accuracy.
\fohcce{} has the highest single-seed accuracy, while CLM-shaped
MLPs have the lowest NLL and ECE. \fohce{} averages
$0.317\pm0.015$ over three seeds, so small accuracy gaps should
be read alongside seed variation.

\begin{table*}[!t]
\centering\small
\setlength{\tabcolsep}{4pt}
\begin{tabular}{lcccc}
\hline
Head (same joint states) & Params & Acc.\ $\uparrow$ & NLL $\downarrow$ & ECE $\downarrow$ \\
\hline
\fohce{} (seed 0) & 1.05M & 0.304 & 2.19 & 0.194 \\
\fohce{} (seeds 0, 1, 2) & 1.05M & $0.317\pm0.015$ & -- & -- \\
\fohcce{} & 5.38M & \textbf{0.331} & 2.08 & 0.118 \\
CLM-shaped MLPs (joint) & 18.9M & 0.327 & \textbf{1.98} & \textbf{0.067} \\
\hline
\end{tabular}
\caption{MMLU-Pro accuracy, NLL, and ECE of heads trained with
one-hot CE on the same joint final-layer states, including the
CLM-shaped MLPs. Seed 0 unless stated; NLL/ECE use raw probabilities.
Bold compares single-seed results.}
\label{tab:gold-heads}
\end{table*}

\subsection{Validation KL by Exit Depth}
\label{app:kl-details}
Later layers reduce validation KL without consistently improving
test accuracy (Tables~\ref{tab:head-kl} and~\ref{tab:depth}). At
layer 24, \fohc{} has higher KL than \foh{} (0.103 versus 0.091)
yet slightly higher accuracy. At layer 32, it has lower KL
(0.034 versus 0.060).

\begin{table*}[!t]
\centering\small
\begin{tabular}{lccccc}
\hline
Readout layer & 16 & 20 & 24 & 28 & 32 \\
\hline
\foh{} & 0.551 & 0.190 & 0.091 & 0.061 & 0.060 \\
\fohc{} & 0.474 & 0.176 & 0.103 & 0.039 & 0.034 \\
\hline
\end{tabular}
\caption{Validation KL ($\downarrow$) by exit depth on 1{,}903 rows
from the training pool (seed 0). Accuracy for these heads is in
Table~\ref{tab:depth}.}
\label{tab:head-kl}
\end{table*}

\subsection{Expanded Training Pool}
\label{app:data-details}
Table~\ref{tab:data} compares the 38k and 123k pools across exit
depths. This single-seed comparison changes sources and validation
splits as well as size, so it does not isolate data quantity.

\begin{table*}[!t]
\centering\small
\begin{tabular}{lcccc}
\hline
 & \multicolumn{2}{c}{\foh{}} & \multicolumn{2}{c}{\fohc{}} \\
Layers run (of 36) & 38k & 123k & 38k & 123k \\
\hline
16 & 0.579 & 0.564 & 0.569 & 0.605 \\
20 & 0.708 & 0.719 & 0.716 & 0.726 \\
24 & 0.747 & 0.749 & 0.750 & 0.751 \\
28 & 0.756 & 0.759 & 0.757 & 0.756 \\
32 & 0.753 & 0.756 & 0.755 & 0.756 \\
\hline
36 + LM head (teacher) & \multicolumn{4}{c}{0.757} \\
\hline
\end{tabular}
\caption{Equal-weight eight-benchmark accuracy with the 38k/123k
pools (seed 0); the teacher is the last row. Table values are rounded;
text differences use unrounded scores.}
\label{tab:data}
\end{table*}

\section{Baselines and Agent Evaluation}
\label{app:baselines}

\subsection{CLM Baselines}
\label{app:clm}
Released CLM uses separately encoded, L2-normalized states and
normalized MLP outputs, with cosine scale capped at 100.

We fine-tune the released CLM heads with the backbone frozen,
using the authors' script: AdamW, batches of 256, peak learning
rate $5\times10^{-4}$, zero weight decay, gradient clipping at 1.0,
and up to 20 epochs with patience 5. The one-cycle schedule warms
up for 10\% of steps. Its 90/10 row split and bidirectional in-batch
InfoNCE differ from our 95/5 question split and per-question CE
controls. Validation accuracy selects the checkpoint, including
the initialization.

All methods use raw
probabilities and top-option confidence for ECE (15 equal-width bins).

\subsection{Teacher-Judge Protocol and Uncertainty}
\label{app:agentic}
The judge uses thinking-disabled Qwen3-8B chat without a system
prompt. Its 8{,}191-token budget reserves the final turn and
assistant prefix, keeps up to 2{,}048 task tokens, and fills the
rest with recent run text. The final user turn is (wrapped for display):
\begin{center}
\footnotesize\ttfamily\frenchspacing
\begin{tabular}{p{\dimexpr\linewidth-2\tabcolsep\relax}}
Above is an AI agent's attempt to resolve the issue.\\
Will the repository's hidden tests pass with the agent's final changes?\\
Answer with yes or no.
\end{tabular}
\end{center}
We combine \texttt{yes}/\texttt{Yes} and \texttt{no}/\texttt{No}
probabilities and renormalize. The student uses the same budget
with shuffled, lettered yes/no options; its answer state is the
last assistant-prefix token after the empty thinking block.

The lettered prompt provides option states for the student.
With this prompt, the full model selects passing runs on 74 tasks
with AUC 0.769, versus 74 and 0.760 for the yes/no judge.
The judge says yes to 94\% of runs, and its top run has probability
above 0.999 in every task.

\foh{} agent-selection heads train for 30 epochs, holding out ten tasks per seed.

Paired 95\% intervals resample
100 test tasks 10{,}000 times, comparing the head's five-seed
mean with the judge on the same tasks (Section~\ref{sec:agentic}).

\section{DeepSWE Evaluation with Separate Embeddings}
\label{app:deepswe}

We evaluate CLM's published DeepSWE held-out split\footnote{Repository:
\href{https://huggingface.co/datasets/tarsur385/deepswe-prm-embeddings-8k}{\nolinkurl{tarsur385/deepswe-prm-embeddings-8k}}.}:
38 tasks and 151 Opus~5 runs, four per task except one with three.
The release supplies separate state/action embeddings and metadata,
but no run text. All heads in Table~\ref{tab:deepswe}, including
\fohce{} and \fohcce{}, use these embeddings; this tests separate
reading, while the missing text prevents joint encoding and
teacher-based KD.

\begin{table*}[!t]
\centering\small
\begin{tabular}{lcc}
\hline
Selector & Tasks solved & Success rate \\
\hline
Random & 28.0 & 73.7\% \\
CLM-shaped, from scratch & 29.6$\pm$1.3 & 77.9\% \\
\fohce{} (read separately) & 30.2$\pm$0.4 & 79.5\% \\
\fohcce{} (read separately) & 29.4$\pm$0.9 & 77.4\% \\
CLM, released DeepSWE head \cite{kwok2026clm} & \textbf{31} & \textbf{81.6\%} \\
\hline
Oracle & 34 & 89.5\% \\
\hline
\end{tabular}
\caption{DeepSWE selection on 38 tasks using released separate
state/action embeddings. From-scratch heads learn one-hot action
CE and report five-seed means $\pm$ sample standard deviations.
Released CLM solves 31 tasks in both its authors' report and our
reproduction. Model names denote head architectures with separate reading.}
\label{tab:deepswe}
\end{table*}

Our heads train from scratch with one-hot action cross-entropy on
passing-run embeddings: 20{,}117 steps for fitting and 2{,}459 for
validation. Each training row has its own action, 15 same-task
actions, and 16 random actions. Test rows use the own action and
31 fixed training-pool references.

We score runs by averaging own-action log-probabilities over the
final 12 steps. The released CLM DeepSWE head uses its fine-tuned checkpoint
and averages cosine scores. Losses and training histories also
differ, so this comparison does not isolate architecture or pretraining.

All runs pass on 21 tasks and fail on four, leaving 13 tasks that
can distinguish selectors. One task changes the success rate by
2.6 percentage points; seed variation reaches 1.3 tasks.
Table~\ref{tab:deepswe} reports each selector's score.

CLM's released corpus spans 24 models and more than 450{,}000 agent steps.
Regenerating it would be costly and would produce different stochastic runs.

\section{Inference Efficiency Protocols}
\label{app:timing}

\paragraph{Native vLLM.}
Table~\ref{tab:efficiency} uses vLLM 0.22.1 and PyTorch 2.11.0
with CUDA 13.0. Each model runs in a fresh process, with BF16
backbone weights, FP32 heads, no CPU offload, and eager execution.
The causal pooling engine runs the four FunDM variants and the full teacher;
the teacher applies its original full-vocabulary LM head to the
answer state and normalizes over valid option letters.
Chunked prefill is disabled; the teacher and FunDM variants also
disable prefix reuse. All models have
capacity for 32{,}768 KV tokens, a 4{,}096-token context limit,
16 active sequences, and a 16{,}384-token batch budget. CE retains
all 36 decoder layers and uses its original prompt and readout.
CLM uses its authors' Engine, embedding cache, and a 64-MiB vector
cache, with a native vLLM encoder. Prefix caching is enabled. The
embedding, vector, and prefix caches start empty for each timed pass. Vector-cache
hits are zero and prefix hits are 0.82\% on this workload; within-pass
deduplication remains enabled. Its GPU footprint includes both
the encoder worker and the head/cache process.

We shuffle the MMLU-Pro rows with seed 0 and take 500 whose FunDM
and CE prompts fit the context limit. After a full untimed pass, we record three timed
passes. First-decision latency uses 50 sequential batch-one requests.
The worker resets CUDA allocator peaks after warmup. The measured
peak includes weights, KV tensors, activations, workspaces, and
reserved slack; it excludes non-PyTorch allocations and the CUDA
context. Tokenization, HTTP, and queueing are outside timing.
The raw measurements and sample IDs are in
\texttt{assets/serving-data.json}.

Before timing, 20 rows are checked against the Hugging Face
feature cache. Head decisions agree on 19--20 rows; released CLM
agrees with its saved Engine reference on all 20. The teacher
agrees on 18 rows under strict argmax and on all 20 when its two
exact reference ties are allowed. Its mean and maximum probability
differences are 0.003 and 0.057. Accuracy tables use the Hugging Face
evaluations; these checks validate the native readout.

For equal-length inputs, 24 of 36 layers save one third of
decoder-layer operations and potential KV storage. These are
architectural estimates; measured memory is the CUDA allocator
footprint on the Spark's shared CPU/GPU memory. Parameter counts
are reported in Table~\ref{tab:depth} and Section~\ref{sec:method}.

\paragraph{PyTorch Throughput.}

We time backbone and head execution in one session on the same
DGX Spark GB10, with Qwen3-8B, BF16 precision, and 500 questions
per benchmark (Table~\ref{tab:throughput}), sampled with seed 0.
All methods use a common PyTorch backbone; this replaces CLM's
vLLM encoder, so the measurements do not test vLLM serving or
prefix caching. Inputs are tokenized before timing and packed by
length with a 16{,}384-token batch budget. CUDA synchronization
brackets each timed pass. After an untimed pass, methods run in
interleaved order for two timed repetitions; we report their median.
Timing uses untrained \foh{} heads with the same architectures as
the fitted heads. The original per-repeat report is archived with the
code as \texttt{assets/timing-data.json}.

CLM's embedding and vector caches start empty for each timed pass.
``Options prepared'' precomputes question-independent option vectors
for all 500 questions; timing covers question encoding and scoring
against those cached vectors. It excludes option preparation and
assumes known candidates, so it does not measure end-to-end cost
for previously unseen options.

\foh{} and the teacher share a prompt, averaging 216 tokens on
MMLU-Pro versus 182 for CE. \fohc{} throughput is within 0.4\% of
\foh{}. \foh{} reaches $1.50\times$ and $1.80\times$ the teacher's
throughput at 24 and 20 layers. On MMLU-Pro, layer 32 gives
15.8 questions/s ($1.13\times$ the teacher). Averaging each
benchmark's throughput ratio over the eight benchmarks gives
$1.21\times$ for full-depth \fohce{} and $1.45\times$ for
24-layer \foh{}, relative to CLM with initially empty caches.

\begin{table*}[!t]
\centering\footnotesize
\setlength{\tabcolsep}{3pt}
\resizebox{\textwidth}{!}{%
\begin{tabular}{llcccccccc}
\hline
Method & Layers & MMLU-Pro & MMLU & OBQA & CSQA & PIQA & HellaSwag & WinoG. & BoolQ \\
\hline
CLM, all texts batched & 36 & 15.2 & 22.4 & 54.3 & 55.3 & 37.1 & 18.3 & 55.0 & 15.5 \\
CLM & 36 & 14.1 & 20.5 & 51.3 & 56.5 & 31.0 & 17.0 & 94.1 & 15.8 \\
CLM, options prepared & 36 & 31.7 & 35.1 & 103.9 & 107.6 & 140.3 & 53.7 & 107.3 & 19.6 \\
\fohce{} & 36 & 16.2 & 25.7 & 66.2 & 84.6 & 37.3 & 18.5 & 93.9 & 19.5 \\
Full model (teacher) & 36 + LM head & 14.0 & 22.1 & 49.8 & 53.1 & 32.2 & 16.6 & 62.4 & 17.2 \\
\hline
\foh{} & 32 & 15.8 & 25.0 & 56.1 & 59.9 & 36.4 & 18.7 & 70.5 & 19.4 \\
\foh{} & 28 & 18.0 & 28.5 & 64.1 & 68.4 & 41.5 & 21.3 & 80.5 & 22.2 \\
\foh{} & 24 & 21.0 & 33.2 & 74.9 & 79.8 & 48.3 & 24.9 & 93.9 & 25.9 \\
\foh{} & 20 & 25.2 & 39.8 & 89.6 & 95.6 & 58.0 & 29.8 & 112.4 & 31.0 \\
\foh{} & 16 & 31.5 & 49.8 & 112.1 & 119.5 & 72.4 & 37.3 & 140.9 & 38.7 \\
\hline
\end{tabular}}
\caption{PyTorch \textbf{throughput} in questions/s on one DGX Spark
(500 questions per benchmark). ``All texts batched'': length-packed
question/options without caching. ``CLM'': embedding/vector caches
start empty; within-pass deduplication is enabled. ``Options prepared'':
option vectors are preloaded, with preparation cost excluded.
\foh{} uses the first $L$ decoder layers.}
\label{tab:throughput}
\end{table*}

\end{document}